\documentclass[journal]{IEEEtran}
\usepackage[caption=false,font=normalsize,labelfont=sf,textfont=sf]{subfig}
\usepackage[pdftex]{graphicx}
\usepackage{stfloats}
\usepackage{rotating}
\usepackage{graphicx}  %Required
\usepackage{epstopdf}
\usepackage{amsmath}
\usepackage{booktabs}
\usepackage{amssymb}
\usepackage{textcomp,booktabs}
\usepackage{multirow}
\usepackage[usenames]{color}
\usepackage{booktabs}
\usepackage{makecell}
\usepackage{mathrsfs}
\usepackage{amsmath}
\usepackage{txfonts}
\usepackage{textcomp,booktabs}
\usepackage{multirow}
\usepackage{amsmath}
\usepackage[usenames]{color}
\usepackage{wrapfig}
\usepackage{bbding}
\usepackage{longtable}
\usepackage{colortbl,booktabs}
\usepackage{verbatim}
\usepackage{caption}
\usepackage{pifont}
\usepackage[colorlinks,linkcolor=blue]{hyperref}
\begin{document}

\title{A Comprehensive Survey on Video Saliency Detection with Auditory Information: the Audio-visual Consistency Perceptual is the Key!}

\author{\IEEEauthorblockN{Chenglizhao Chen$^{1}$~~~~Mengke Song$^{1}$~~~~Wenfeng Song$^{3}$~~~~Li Guo$^{1\dag}$~~~~Muwei Jian$^{2}$\\
		\IEEEauthorblockA{$^{1}$China University of Petroleum (East China)~~~~~~~~$^{2}$Shandong University of Finance and Economics\\
$^{3}$Beijing Information Science and Technology University\\
		Code \& Data: \url{https://github.com/songsook/SCDL}\vspace{-0.8cm}
			\thanks{\dag Corresponding author: Li Guo (ally\_kwok@163.com)}
			\thanks{The first two authors contribute equally to this paper.}
		}}}

\markboth{IEEE Transactions on Circuits and Systems for Video Technology, VOL.XX, NO.XX, XXX.XXXX}%
{Shell \MakeLowercase{\textit{et al.}}: Bare Demo of IEEEtran.cls for Journals}

\maketitle

%\IEEEtitleabstractindextext{
\begin{abstract}
Video saliency detection (VSD) aims at fast locating the most attractive objects/things/patterns in a given video clip. Existing VSD-related works have mainly relied on the visual system but paid less attention to the audio aspect, while, actually, our audio system is the most vital complementary part to our visual system.
Also, audio-visual saliency detection (AVSD), one of the most representative research topics for mimicking human perceptual mechanisms, is currently in its infancy, and none of the existing survey papers have touched on it, especially from the perspective of saliency detection.
Thus, the ultimate goal of this paper is to provide an extensive review to bridge the gap between audio-visual fusion and saliency detection.
In addition, as another highlight of this review, we have provided a deep insight into key factors which could directly determine the performances of AVSD deep models, and we claim that the audio-visual consistency degree (AVC) --- a long-overlooked issue, can directly influence the effectiveness of using audio to benefit its visual counterpart when performing saliency detection.
Moreover, in order to make the AVC issue more practical and valuable for future followers, we have newly equipped almost all existing publicly available AVSD datasets with additional frame-wise AVC labels.
Based on these upgraded datasets, we have conducted extensive quantitative evaluations to ground our claim on the importance of AVC in the AVSD task.
In a word, both our ideas and new sets serve as a convenient platform with preliminaries and guidelines, all of which are very potential to facilitate future works in promoting state-of-the-art (SOTA) performance further.
\end{abstract}

\begin{IEEEkeywords}
audio-visual Fusion; Video Salieny Detection; semantical consistency.
\end{IEEEkeywords}
%}
\maketitle
%\IEEEdisplaynontitleabstractindextext
\IEEEpeerreviewmaketitle

%\begin{figure*}[t]
%	\centering
%	\includegraphics[width=1\linewidth]{Motivation.png}
%    \vspace{-0.4cm}
%	\caption{The major difference between the current SOTA models and our approach. As seen in subfigure-A, the conventional VSOD models usually take several consecutive video frames as input; thus, their VSOD methodology follows the short-term manner, where the current saliency decision is only derived on the current consecutive frames. In sharp contrast, our approach is based on object-level clustering, where all object proposals belonging to different frames (including the beyond-scope frames) are simultaneously available when making saliency prediction, and this is a typical long-term manner, where, compared with the short-term manner, the long-term methodology can be more robust when the salient objects have large appearances or movement changes. We use subfigure-B to visualize the salient object proposal mining process.}
%	\label{fig:Motivation}
%\end{figure*}

\section{Introduction}

We humans tend to be attracted by specific things, and this mechanism has its basic principle in general. But, outwardly, it could vary from different people and scenes, and, directly or indirectly, such differences are usually caused by either personality and individual differences or the exact environment~\cite{Chen-TIP21-Dep,Chen-TIP20-Imp,Ma-TIP21-Ret}.
For example, in an open wild, we may get attracted by a fantastic nature scene view and pay less attention to artificial subjects. However, things go differently in a downtown area, where magnificent artificial buildings could keep drawing our attentions. Also, our attention could get shifted to ``rare'' elements --- patterns that are anomalies for their nearby surroundings, and we have an academic name for all these objects/things/patterns attracting our attention --- saliency.

In general, the saliency-related research activities~\cite{Jian-TCYB-IS} should come with a specific venue, \emph{e.g.}, the visual saliency, which aims at segmenting the most eye-attracting objects or regions in a given scene.
And the scenes are usually ``expressed'' in the form of images or videos.
Since video data is the main course of this survey, we shall omit image-based saliency works.

\begin{figure}[!t]
	\centering
	\includegraphics[width=1\linewidth]{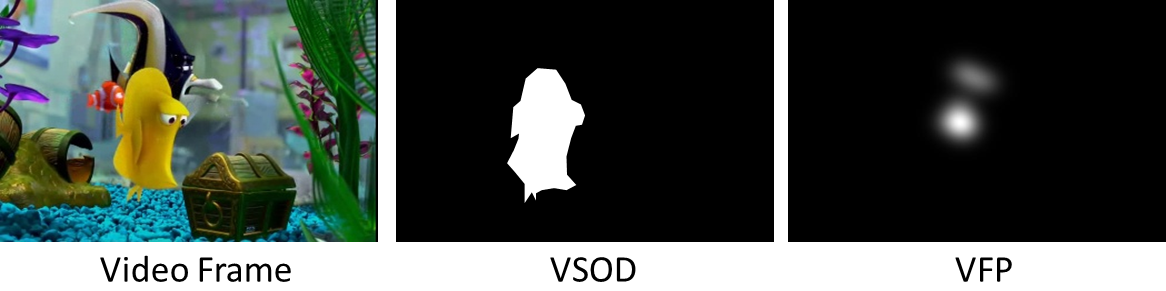}
    \vspace{-0.5cm}
	\caption{The difference between \textbf{v}ideo \textbf{s}alient \textbf{o}bject \textbf{d}etection (VSOD) and \textbf{v}ideo \textbf{f}ixation \textbf{p}rediction (VFP). The GTs for VSOD (the 2nd column) are the \emph{human-annotated} object-level binary masks, while the GTs used in VFP (the 3rd column) are the human-eye fixations (without clear object boundary) recorded \emph{automatically} by using eye-trackers.}
	%\vspace{-0.6cm}
	\label{fig:VSODVFP}
\end{figure}

The current visual saliency detection research field can be roughly divided into two groups, \emph{i.e.}, \textbf{v}ideo \textbf{s}alient \textbf{o}bject \textbf{d}etection (VSOD) and \textbf{v}ideo \textbf{f}ixation \textbf{p}rediction (VFP).
The basic methodologies of VSOD and VFP are almost the same, where the existing hand-crafted methods~\cite{Chen-TMM18-BIL,Chen-LSP18-BU,Chen-TIP17-STF,Chen-PR16-NIS} mainly follow either top-down or bottom-up rationale.
After entering the deep learning era, most of the existing works~\cite{sj2021tcsvt,chen2019improved,li2019accurate,Jian-TCYB-INS} have adopted the end-to-end encoder-decoder network architecture, which, generally, belongs to the typical top-down category. Hence the difference between VSOD and VFP is the exact training ground truth data and training loss functions.
For a better understanding, Fig.~\ref{fig:VSODVFP} has demonstrated such a difference.

\begin{figure*}[!t]
	\centering
	\includegraphics[width=1\linewidth]{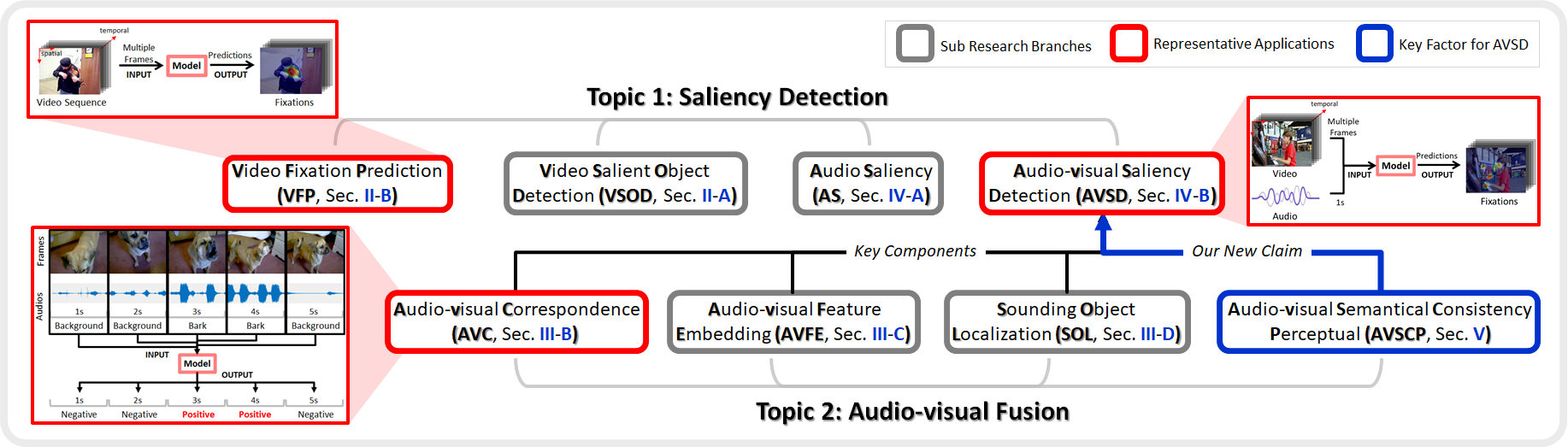}
	\caption{The structure of our review, which covers two significant topics: 1) Video Saliency Detection and 2) Audio-Visual Multi-Modality Fusion. W.r.t., the most representative applications, we have highlighted them with red rectangular boxes. Also, we have newly argued that the audio-visual semantical consistency perceptual (highlighted by the blue box) is the key factor in determining the AVSD performance.}
	\label{fig:Structure}
\end{figure*}

Though our visual system is one of the most important venues for us to perceive the environment that we're in, our auditory system also plays an important role. For example, our attention could fast shift to a sounding object, showing that our auditory system can really complement our visual system.
Despite being complementary in general, these two venues have completely different perceptual mechanisms.

The visual venue is very informative yet with rather limited sensing scope (because of the limited \textbf{f}ield \textbf{o}f \textbf{v}ision, FOV). In contrast, the auditory venue is less informative, yet its sensing scope is clearly dead-angle-free.
Besides, different from the visual saliency research field, a quite mature topic, audio-related saliency is in its infancy.
Moreover, different from the visual saliency, a single modality task with abundant accessible training data, the available training data for the \textbf{a}udio-\textbf{v}isual \textbf{s}aliency \textbf{d}etection (AVSD) is in a critical shortage\footnote{The widely-used VSD and VFP training datasets comprise totally 1.6K video clips, while the available data for AVSD is only 0.2K clips, not to mention the fact that the AVSD task is more challenge than VSD and VFP, and thus is more data-hungry.}, which has resulted in a clear performance bottleneck, especially in this deep learning era.

%\begin{figure}[!t]
%	\centering
%	\includegraphics[width=0.8\linewidth]{Overview.png}
%	\caption{Demonstrations of three most representative applications towards the key topics (\emph{video saliency detection} and \emph{audio-visual multi modality fusion}) of our review.}
%	\vspace{-0.2em}
%	\label{fig:overview}
%\end{figure}

\begin{table}[!t]
	\caption{Illustration of the main differences between the existing reviews and ours.}
	\vspace{-0.2cm}
	\begin{center}
		\renewcommand{\arraystretch}{1.1}{
			\setlength{\tabcolsep}{20pt}{
				\resizebox{1.0\linewidth}{!}{
\Huge
					\begin{tabular}{r||c|c|c}
						\Xhline{2.0pt}
						\hline
						Reviews & Year &Publication & Contents\\
						\hline
						\hline			
												
						Katsaggelos \emph{et.al}~\cite{Katsaggelos-JPROC15-review} & 2015& P-IEEE & Audio-visual Fusion  \\
						Baltrusaitis \emph{et.al}~\cite{Baltrusaitis-TPAMI18-review} & 2018&T-PAMI & Multi-modality Machine Learning \\
						Cong \emph{et.al}~\cite{Cong-TCSVT18-review} & 2018&T-CSVT & RGBD/Video/Co-saliency Detection\\
						Wang \emph{et.al}~\cite{Wang-TPAMI19-review} & 2019& T-PAMI &Video Saliency Detection  \\
						Zhu \emph{et.al}~\cite{Zhu-IJAC21-review} & 2021 & IJAC & Audio-visual Localization/Correspondence\\
						Chen \emph{et.al} (Ours) & 2022 & T-CSVT & Audio-visual Saliency Detection \\
						\hline

						\Xhline{2.0pt}
		\end{tabular}}}}
		\label{tab:reviews}%
		\vspace{-0.6cm}
	\end{center}
\end{table}%

Meanwhile, we have noticed that there exist massive researches~\cite{OTS-ECCV18,SOL-CAM2-TPAMI21,SOL-CAM3-ECCV18} regarding the visual and auditory fusion, and their main interests are usually focusing on the multimedia applications, \emph{e.g.}, multi-modality information processing, filtering, and understanding, and these works are rarely intercrossed with the saliency detection research field.
Though some of the existing fusion methods proposed in previous literatures~\cite{AVCCL3-ARXIV21,AVCRepresent1-ICASSP21} can really inspire and help the network design toward the saliency detection task, none of them has covered both saliency detection and audio-visual fusion.
Thus, as shown in Fig.~\ref{fig:Structure}, this review mainly focuses on two topics, and we choose three concrete research fields as the main courses, \emph{i.e.}, \textbf{v}ideo \textbf{s}aliency \textbf{d}etection (VSD), \textbf{a}udio-\textbf{v}isual \textbf{c}orrespondence (AVC), and \textbf{a}udio-\textbf{v}isual \textbf{s}aliency \textbf{d}etection (AVSD). Also, the differences between several existing reviews on audio-visual representation learning and ours have been illustrated in Table~\ref{tab:reviews}.

Despite providing an extensive review, we have noticed that the \textbf{a}udio-\textbf{v}isual \textbf{c}onsistency (AVC) between audio and visual, a representative task considered in the multimedia research field~\cite{Chen-TASLP21-TSC,Chang-TAFFC21-esc,Han-ICASSP20-CDCU}, is the key factor to determining the overall performance of AVSD, while its importance has long been overlooked by our AVSD research field.
To verify our claim, we have newly labeled all publicly available AVSD clips frame-by-frame, and conducted massive quantitative experiments with them.
This new finding is very potential to benefit our audio-visual saliency detection research field in the near future.

In a brief summary, significant highlights and contributions of this review include the following aspects:
\begin{itemize}
	\item
	This review is the first attempt to bridge the gap between saliency detection and audio-visual fusion;
	\item
    We have extensively included the most recent deep learning-based works, making this review fresh in essence and capable of helping new hands to join this new research topic;
    \item
    We have noticed one critical factor --- the semantical consistency degree, which has been well studied by the multimedia research field while being completely omitted by our AVSD research field, could significantly influence the AVSD performance;
    \item
    For all widely-used existing AVSD datasets, we have newly equipped them with frame-wise semantical consistency degree labels, which could be very potential to benefit our research community.
\end{itemize}

\section{Video Saliency Detection}
In the video saliency detection research field, there exist two main research branches, including the \textbf{v}ideo \textbf{s}alient \textbf{o}bject \textbf{d}etection (VSOD)~\cite{chen17ST,chen2018bilevel,Wu-ICA3PP22-SA,Wang-TCYB22-JDLC,Li-AAAI22-YOIO,Lan-AAAI22-SN} and the \textbf{v}ideo \textbf{f}ixation \textbf{p}rediction (VFP)~\cite{Zhang-ICCV21-DTN,Lai-TIP20}.
The major differences between VSOD and VFP lie in two aspects: training GTs and loss functions.

As shown in Fig.~\ref{fig:VSODVFP}, the GTs used for the VSOD task are binary masks, where all salient objects have been well annotated/segmented by humans. While the GTs used in the VFP task are human-eye fixations (\emph{i.e.}, individual pixel-wise coordinates) collected by eye-trackers directly, representing raw image regions that humans would pay attention to.
In a word, GTs for the VSOD task are object-aware, while GTs for the VFP task are scattering locations.

Also, the widely-used loss function in the VSOD task is the cross-entropy loss, while the \textbf{k}ullback-\textbf{l}eibler (KL) divergence is the most frequently used one in the VFP task.
In the following two subsections, we will review these two research branches respectively.

\subsection{\textbf{V}ideo \textbf{S}alient \textbf{O}bject \textbf{D}etection (VSOD)}

As can be seen in Fig.~\ref{fig:VSODVFP}, the major difference between the deep learning-based VSOD and VFP tasks is the different training ground truth data, \emph{i.e.}, the VFP task simply aims at simulating human eye-fixations, while the VSOD is more biased towards the object aspect, which can be treated as a combination of object segmentation and object localization~\cite{li2019accurate,li2021tcsvt,sj2021tcsvt}, where the rationale of the localization process is similar to that of the VFP task, and all those non-salient objects are filtered by this process.

\begin{table}[!t]
	\caption{Strengths and weaknesses comparison between Optical Flow~\cite{OpticalFlow_CLiu}, LSTM~\cite{LSTM}, ConvLSTM~\cite{Shi-NIPS15-ConvLSTM} and 3D Convolution~\cite{Tran-ARXIV14} towards temporal sensing. \{M. S.\}: multi-scale, \{O. F.\}: optical flow, \{P. B.\}: performance bottleneck, \{F. ST. I.\}: full spatiotemporal interaction; {\color{red}\XSolidBrush}: \emph{without}, {\color{green}\ding{51}}: \emph{with}.}
	\vspace{-0.2cm}
	\begin{center}
		\renewcommand{\arraystretch}{1.2}{
			\setlength{\tabcolsep}{1.5pt}{
				\resizebox{1.0\linewidth}{!}{
					\begin{tabular}{r||c|c|c|c|c|c|c}
						\Xhline{1.0pt}
						\hline
						Methods &Implement&M. S.&Computation&Network&O. F. &P. B.&F. ST. I.\\
						%& \\
						\hline
						\hline			
						Optical Flow~\cite{OpticalFlow_CLiu} &\emph{easy} &\emph{\color{green}\ding{51}}&\emph{expensive}&\emph{light}&\color{green}\ding{51}&\color{green}\ding{51}&\color{red}\XSolidBrush \\
						LSTM~\cite{LSTM} &\emph{hard} &\emph{\color{red}\XSolidBrush}&\emph{expensive}&\emph{heavy}&\color{red}\XSolidBrush&\color{green}\ding{51}&\color{red}\XSolidBrush \\
						ConvLSTM~\cite{Shi-NIPS15-ConvLSTM} &\emph{hard} &\emph{\color{red}\XSolidBrush}&\emph{expensive}&\emph{heavy}&\color{red}\XSolidBrush&\color{green}\ding{51}&\color{red}\XSolidBrush\\
						3D Convolution~\cite{Tran-ARXIV14} &\emph{easy} &\emph{\color{green}\ding{51}}&\emph{cheap}&\emph{light}&\color{red}\XSolidBrush&\color{red}\XSolidBrush&\color{green}\ding{51}\\

						\hline
						
						\Xhline{1.0pt}
		\end{tabular}}}}
		\label{comparison}
		\vspace{-0.6cm}
	\end{center}
\end{table}%

Despite using different GTs, there also exist multiple other distinguishing differences:

1) The VSOD task should additionally consider detections' integrity, \emph{i.e.}, the detected salient regions should precisely comprise the entire salient object with all its subparts included. However, the VFP task aims at the simulation of the human eye's fixation, thus the detected results are not required to highlight the entire object.

2) The widely-used VSOD scenario could be full automatic video segmentation. In this application, the saliency ranks of different objects tend to stay unchanged for a long period of time. However, the human eye's fixations are usually scattered locations, which are rather weak in indicating those corresponding objects. In other words, fixations usually shift between objects.

Besides, due to the differences mentioned above, the loss functions adopted by VSOD and VFP are also different, where the VSOD task is mainly using the cross-entropy loss, while the VFP task usually prefers the \textbf{k}ullback-\textbf{l}eibler (KL) divergence, linear \textbf{c}orrelation \textbf{c}oefficient (CC) loss, \textbf{n}ormalized \textbf{s}canpath \textbf{s}aliency (NSS) loss, and \textbf{sim}ilarity (SIM) loss, where all these losses are designed for measuring the consistency degree between the predicted scattering fixations and the real human eye fixations.

The existing \textbf{s}tate-\textbf{o}f-\textbf{t}he-\textbf{a}rt (SOTA) VSOD models~\cite{Lee-AAAI22-Ite,Xu-AAAI22-Rel,Chen-WACV22-CFAM,Lu-ARXIV22-Dep,Zhao-CVPR21-WS,Tang-ARXIV21-ALGR,Jiao-ICIP21-GAT,Zhao-ACMMM21-MSF} can be divided into two groups according to their network designs: 1) the bi-stream-based methods~\cite{li2019motion_mga,chen2019improved,ren20TENet,Zhang-ICCV21-DCSF,Ji-ICCV21-FDS}, and 2) the single-stream-based ones~\cite{song2018pyramid_pdbm,fan2019shifting_ssav,gupyramid_pcsa,Wang-TIP21}.

The bi-stream-based models usually consist of two sub-branches, one for the motion saliency clues, whose input focuses on the temporal information (\emph{e.g.}, optical flow data); another branch is the conventional color branch, which could be any off-the-shelf image salient object detection deep model.
Note that, the network architectures of these two branches could be the same, and the only difference is their training input, \emph{i.e.}, optical flow result \emph{vs.} color image.

The single-stream-based methods have abandoned the individual temporal computation, \emph{e.g.}, the time-consuming optical flow~\cite{OpticalFlow_CLiu}. Instead, it takes multiple frames as input each time, and then uses either LSTM~\cite{LSTM}, ConvLSTM~\cite{Shi-NIPS15-ConvLSTM} or 3D convolution~\cite{Tran-ARXIV14} to sense temporal information. Detailed comparison results of these methods are shown in Table~\ref{comparison}. Compared with the bi-stream-based methods, this type of work has a significant advantage, \emph{i.e.}, it could be 10 times faster in computation, because the individual temporal information computation is the major efficiency bottleneck for the bi-stream-based approaches.
More details regarding this issue can be found in~\cite{Wang-TIP21}.

\subsection{\textbf{V}ideo \textbf{F}ixation \textbf{P}rediction (VFP)}
Different from the VSOD task, which uses well-annotated object-wise binary masks as training objectives, the training GTs for the VFP task are scattered human-eye fixation locations collected by the eye tracker (\emph{e.g.}, Tobbi, EyeLink, Smart Eye, and GazeTech).
The earliest deep learning-based VFP approaches~\cite{Bak-TMM18} followed the bi-stream structure, which clearly belongs to the multi-task rationale, where one stream handles the fixation predictions in the spatial domain, and another stream focuses on the fixation predictions over the temporal scale.
Thus, the key problem of the bi-stream-based VFP models is how to achieve the fusion balance between its sub-streams.
Clearly, this methodology is quite similar to that of the bi-stream-based VSOD approaches, while tiny differences would be the different loss functions and GTs.

\begin{figure}[!t]
\centering
\includegraphics[width=1\linewidth]{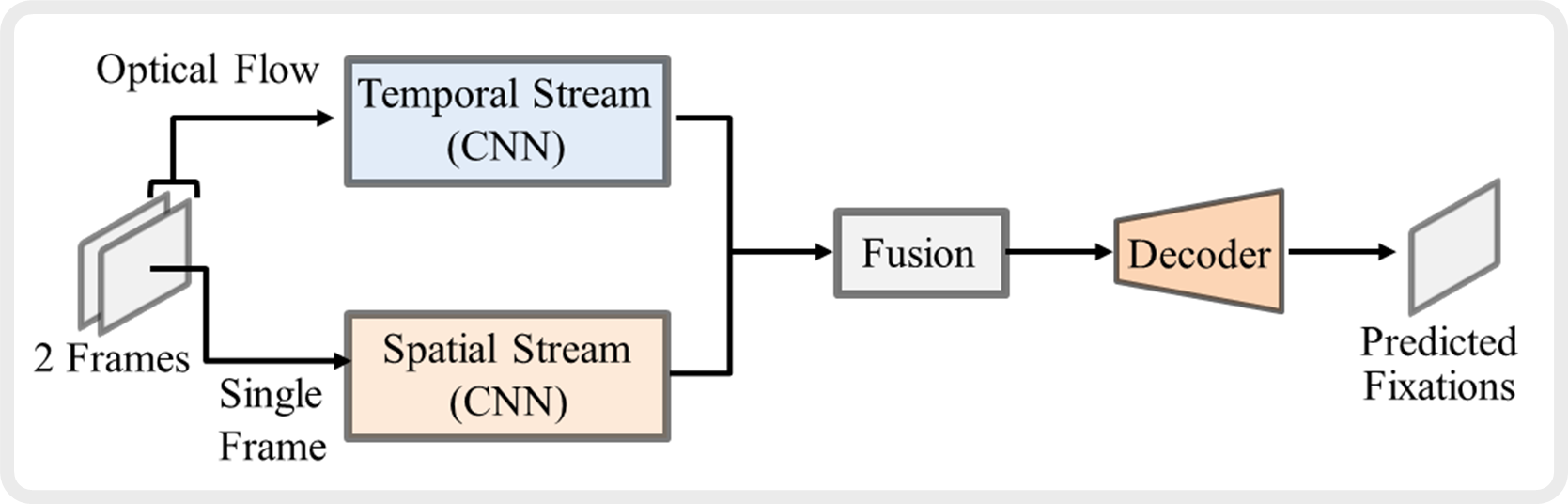}
\caption{Method pipeline of the optical flow-based bi-stream approaches, which mainly contains temppral stream and spatial stream followed by a fusion module and a decoder.}
\vspace{-0.2em}
\label{fig:bisofpipe}
\end{figure}

\vspace{0.2cm}
\textbf{Optical Flow-based Approaches}. The primary way for the bi-stream VFP models to sense temporal information is to take the optical flow results as the models' input. As shown in Fig.~\ref{fig:bisofpipe}, almost all existing bi-stream VFP models have adopted the optical flow (\emph{e.g.}, the most representative conventional one~\cite{OpticalFlow_CLiu} and the deep learning-based ones, such as FlowNet~\cite{FlowNet,FlowNet2}) as the temporal sub-stream to sensing temporal information. Here we just name a few most representative ones.

In~\cite{Lai-TIP20}, Lai~\emph{et al.} have made two key innovations: 1) a novel way for performing early fusion between spatial and temporal feature backbones, and 2) the \textbf{conv}olutional \textbf{G}ated \textbf{R}ecurrent \textbf{U}nit (convGRU) has been firstly applied for learning the temporal attention transitions across time, which is able to make the predicted video fixation maps temporally smooth.
The major highlight of the fusion scheme is that the deep features, obtained by the spatial and temporal feature backbones respectively, are connected densely via residual attention mechanism in a multi-scale way.
Specifically, the exact fusion operation has clearly biased towards the spatial information, where the deep features from the temporal backbone are only served as auxiliary stimuli.
As a variant of the classic LSTM, the proposed convGRU has two advantages: 1) simpler in network design, and 2) slight performance improvement (less than 0.5\%).

Following the bi-stream structure~\cite{Zhang-ICCV21-DTN} also, Zhang~\emph{et al.}~\cite{Zhang-TIP21} have devised a novel fusion scheme.
The key idea of the proposed fusion is to perform a selective combination of spatial and temporal information. The channel-wise attention has been used as the indicator to guide the selection process, and the rationale is that only those deep features with strong feature responses would be able to benefit the detection task. In addition, the authors have devised a novel strategy which takes the spatial position of the salient objects in previous consecutive frames as the additional input, aiming at facilitating the estimation of temporal saliency by shrinking the problem domain. Consequently, the network's output could stay consistent (smooth) over the temporal direction.

In summary, the major advantage of the optical flow-based approaches is the strong temporal sensing ability, while the disadvantage is also clear, \emph{i.e.}, the optical flow computation process takes some additional time. Also, the exact fusion strategy determines the overall performance directly, because the spatial and temporal streams are less interactive actually.

\begin{figure}[!t]
\centering
\includegraphics[width=1\linewidth]{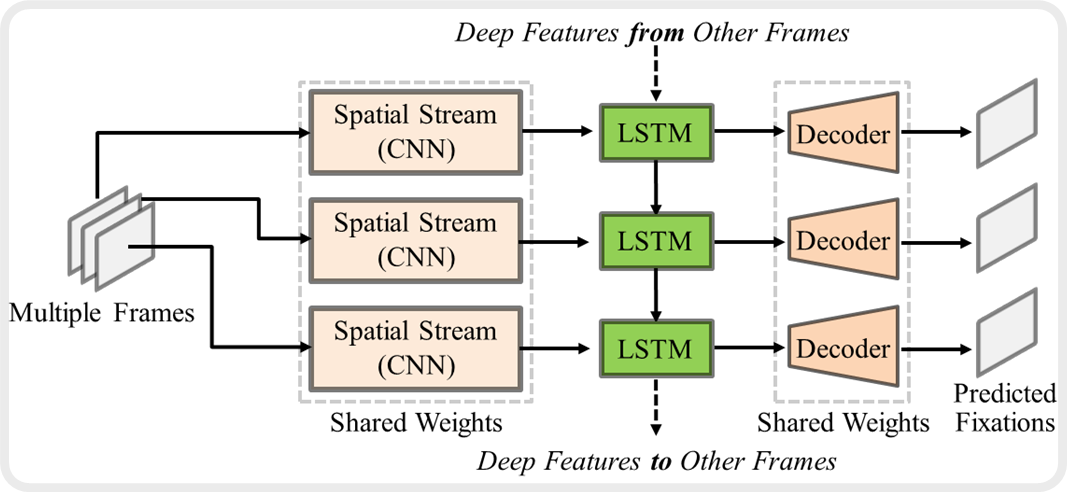}
\caption{Method pipeline of the \textbf{l}ong \textbf{s}hort-\textbf{t}erm \textbf{m}emory- (LSTM-) based approaches which usually follow the single-stream methodology.}
\vspace{-0.2em}
\label{fig:lstmpipe}
\end{figure}

\vspace{0.2cm}
\textbf{Long Short-Term Memory-based Approaches}.
Actually, most of the existing \textbf{s}tate-\textbf{o}f-\textbf{t}he-\textbf{a}rt (SOTA) VFP approaches have adopted the \textbf{l}ong \textbf{s}hort-\textbf{t}erm \textbf{m}emory (LSTM) to sense temporal information. Compared with the optical flow-based ones, the LSTM-based approaches usually follow the single-stream methodology.
As can be seen in Fig.~\ref{fig:lstmpipe}, this type of approaches usually adopts the \textbf{c}onvolutional \textbf{n}eural \textbf{n}etworks (CNN) to compute spatial deep features for each single frame. Then, in order to sense temporal information, all deep features computed individually via CNN are fed into the input gate of LSTM. Finally, a decoder is applied to produce the pixel-wise fixation prediction.

In~\cite{Wang-CVPR18}, Wang~\emph{et al.} have completely followed the structure demonstrated in Fig.~\ref{fig:lstmpipe}. However, some modifications have been made in the spatial stream, including: 1) several residual layers were used to compensate for the loss of receptive field caused by removing the last two pooling layers of the VGG16 feature backbone; 2) the spatial attention mechanism was applied to the spatial-stream for facilitating the network training, where the static fixation GTs could be used as the attentions helping the network's training (\emph{i.e.}, the dynamic saliency), which could be able to alleviate the demand of large scale of costly video fixation GTs.

Similar to~\cite{Wang-CVPR18}, Linardos~\emph{et al.} in~\cite{Linardos-ARXIV19} have placed the LSTM in the middle stage of a typical encoder-decoder CNN. The LSTM collects the output of the encoder, and then its output, representing the spatiotemporal information, is fed to the decoder to formulate the fixation prediction.
The major highlight of this work is the proposed recurrent mechanism, where the LSTM's output is used as an intra-attention to enhance the input data flow. Consequently, the network's ability to sense temporal information gets improved significantly.

To further enhance the sensing ability of temporal information, Chen~\emph{et al.} in~\cite{Chen-PR21} have taken 3 frames as the network's input each time. Then, the deep features respectively computed from these frames are combined as the input of LSTM.
Compared with the conventional LSTM-based approaches which take only 1 frame as input each time, this method has considered 3 frames, and thus its temporal sensing ability could, of course, get enhanced.
Since the spatial displacement occurs along the time scale, deep features computed from consecutive video frames are usually misaligned, which could confuse the subsequent learning process, blurring the final prediction results.
To solve this problem, the authors have resorted to deformable convolution --- an off-the-shelf tool that could dynamically learn the spatial positions of the adopted convolutional kernels.
By using the deformable convolution, the deep features before inputting into the LSTM are aligned.

It is worth mentioning that the LSTM can also be replaced by other networks which have the ability to sense temporal information.
For example, Droste~\emph{et al.} in~\cite{Droste-ECCV20} have adopted the \textbf{r}ecurrent \textbf{n}eural \textbf{n}etwork (RNN), the early prototype of the LSTM, for sensing temporal information, where the RNN is placed between the encoder and decoder, sharing a similar overall network structure to that of the \cite{Chen-PR21}.

Apart from the single-stream LSTM-based approaches mentioned above, there also exist several works~\cite{Jiang-ECCV18} following the bi-stream methodology, where the spatial and temporal information interact with each other as an early fusion. Jiang~\emph{et al.} in~\cite{Jiang-ECCV18} followed the typical bi-stream structure, in which a pruned structure of YOLO is applied as the subnet for sensing the spatial information, and the temporal stream is a pruned FlowNet~\cite{FlowNet}. The multi-scale deep features provided by the spatial stream are collected via the concatenation and batch normalization operations, formulating a coarse localization mask to compress those clearly non-salient backgrounds in the deep spatial features. Meanwhile, the deep features of the temporal stream are also assembled in a way identical to that used in the spatial stream. Finally, both the multi-scale deep features assembled individually from the spatial and temporal streams are concatenated to be fed to an LSTM.

Similar to the early fusion adopted in \cite{Jiang-ECCV18}, Wu~\emph{et al.} in~\cite{Wu-AAAI20} have committed one modification to enhance the temporal sensing ability: the inter-frame correlations are explored by performing the simple dot-product operator along the channel dimension.
Besides, the authors have adopted the spatial attention-based shuffle operation to enhance the spatial stream, where the multi-level deep features are combined and later shuffled.
Both these strong spatial deep features and cross-frame correlation features will be fed into a variant version of the LSTM, named the correlation-based ConvLSTM, where the input gate has been modified to an addition operation-based feature fusion, thus it could be able to simultaneously take two different sources as its input.

In a brief summary, the major advantage of the LSTM-based approaches is its faster computational speed than the optical flow-based methods. However, some of the most recent works~\cite{Wang-TIP21} have argued that the nature of the LSTM might not be sensing the temporal information but constraining its individual inputs to stay partially consistent and thus could be capable of eliminating the intermittent external disturbances. In a word, the LSTM is clearly inferior to the optical flow in sensing temporal information.

\begin{figure}[!t]
\centering
\includegraphics[width=1\linewidth]{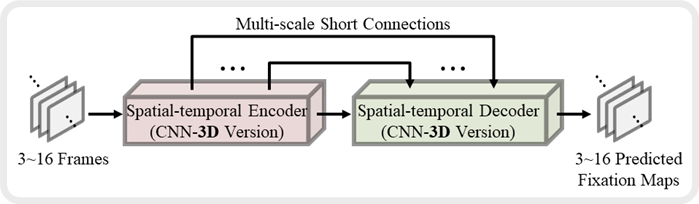}
\caption{Method pipeline of the 3D convolution-based approaches, and the major highlight of these approaches is their capability of sensing both spatial and temporal information in a cubic way.}
\vspace{-0.2em}
\label{fig:3Dpipe}
\end{figure}

\vspace{0.2cm}
\textbf{3D Convolution-based Approaches}.
Compared with the widely-used 2D convolution that can only sense spatial information, 3D convolution is capable of sensing both spatial and temporal information in a cubic way. As been discussed in~\cite{Wang-TIP21}, 3D convolution is generally inferior to its competitors (\emph{e.g.}, LSTM~\cite{LSTM} and optical flow~\cite{Ilg-CVPR17-Flo}) in sensing temporal information, but it still has several unique advantages, \emph{i.e.}, fast computation and good compatibility.
Also, to the best of our knowledge, 3D convolution-based VFP models~\cite{Zou-PRL22-STA3D} are generally leading the SOTA performance, and the overall method pipeline of this type of approach has been provided in Fig.~\ref{fig:3Dpipe}. We shall review several most representatives here.

Min~\emph{et al.} in \cite{Min-ICCV19} have directly applied the 3D convolution to the conventional 2D encoder-decoder architecture, where the exact implementation is straightforward, \emph{i.e.}, all 2D convolutions are replaced by 3D versions.
Though the newly applied 3D convolution is capable of providing some temporal information, there exists one critical problem in the decoder. That is, the widely-used unpooling operation cannot provide the exact spatial locations over the temporal scale, limiting its decoder's performance. To alleviate it, the authors have devised an auxiliary pooling scheme, whose key rationale is to record all spatial, temporal, and channel locations when performing pooling operations. Therefore, the unpooling operations in the decoder layers can re-use the reserved locations eventually.

Recently, Bellitto~\emph{et al.} in~\cite{Bellitto-ARXIV21} have followed the 3D encoder-decoder network structure for the VFP task. The highlight of this approach is the newly proposed decoder, where two new concepts have been considered.
To handle the domain shift problem, each side output of the encoder is assigned to an unsupervised binary classifier, whose primary objective is to follow the adversarial training that minimizes the gap between features respectively learned from the source and target domain.
Besides, for each layer in the decoder, multiple domain-specific priors are dynamically learned, and incorporated to make the network domain-specific, and this strategy could significantly improve quantitative scores further.

\begin{figure*}[!t]
	\centering
	\includegraphics[width=1\linewidth]{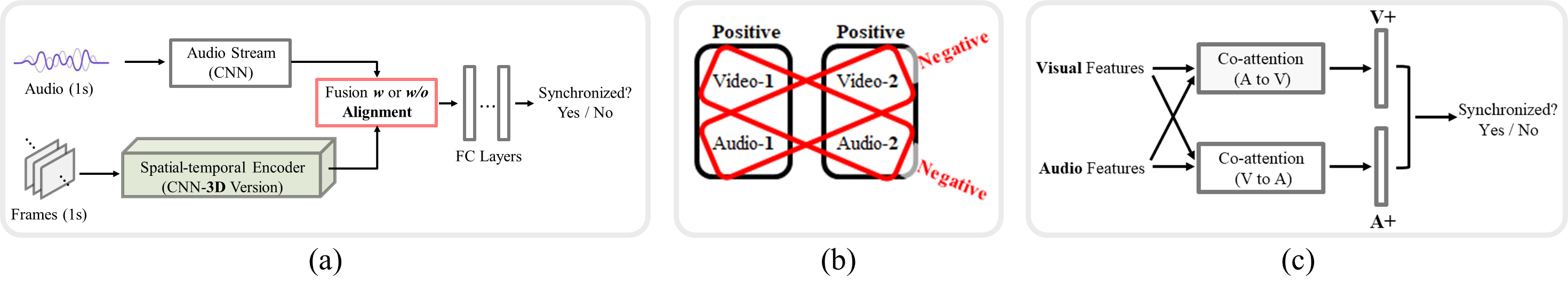}
	\caption{(a) The widely-used network architecture for the \textbf{a}udio-\textbf{v}isual \textbf{c}orrespondence (AVC) task which is a typical binary classification, where the key is how to align and fuse the audio and visual streams. (b) The widely-used \textbf{a}udio-\textbf{v}isual \textbf{c}orrespondence (AVC) training data formulation. The synchronized video and audio pairs are set to positive, whereas the unsynchronized video and audio pairs are set to negative. (c) Co-attention-based audio and visual feature fusion, where the outputs of the co-attention operation can be regarded as the upgraded versions, \emph{i.e.}, $\rm A+$ means upgraded audio features and $\rm V+$ denotes upgraded visual features, where all those clearly unsynchronized information can be effectively excluded.}
	\vspace{-0.8em}
	\label{fig:AVCpipe}
\end{figure*}

\section{Audio-Visual Multi-Modality Fusion}
Different from the visual signal which usually determines human attention directly, the audio signal is usually the auxiliaries, influencing human attention in a subtle way.
For example, our attention can be easily attracted by a sounding object, \emph{e.g.}, the sound of a dropping box hitting the floor.
However, some audio signals are also completely helpless in drawing our attention, \emph{e.g.}, background music.
Thus, since the human visual field has blind spots, the audio signal, whose perception scope is almost $360^{\rm o}$, should be appropriately used to complement visual in practice.
With the development of deep learning techniques, more and more research attentions have been paid to how to combine/fuse audio and visual for vision-related tasks, \emph{e.g.}, sounding object localization~\cite{OTS-ECCV18}, audio-visual synchronization~\cite{AVS-MM20}, object tracking~\cite{AVTracking-ICCV19}, and saliency detection~\cite{WGT-CVPR21}.
Though the primary focus of this review is on saliency detection, we shall still review several most representative audio-visual-related tasks~\cite{Sad-IJST22-CM,Ma-ASB22-DA,Qian-TMM22-AVT} in advance, because these fusion-related arts can be directly referred and get a deep insight towards our audio-visual saliency detection. For a better reading, we propose to introduce three most representative tasks here, including \textbf{a}udio-\textbf{v}isual \textbf{c}orrespondence (AVC), \textbf{a}udio-\textbf{v}isual \textbf{l}ocalization (AVL), and \textbf{f}ace and \textbf{a}udio \textbf{m}atching (FAM).

\subsection{Preliminaries on Audio Feature Representation}
\begin{figure}[!t]
	\centering
	\includegraphics[width=1\linewidth]{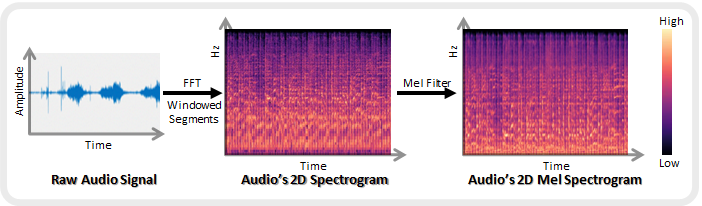}
	\caption{Audio feature computation details. First, the raw 1D audio signal is transformed to a 2D spectrogram by \textbf{f}ast \textbf{F}ourier \textbf{t}ransform (FFT), thus the existing popular backbones (\emph{e.g.}, VggNet~\cite{vgg} or ResNet~\cite{res}) can be used. Next, the Mel filter is utilized to make the 2D spectrogram more discriminative and sensitive to our human auditory system.}
	\vspace{-0.8em}
	\label{fig:AudioPro}
\end{figure}

Given any 1-dimensional raw audio data, we can directly input it to an off-the-shelf feature backbone (\emph{e.g.}, SoundNet~\cite{soundnet} or VggSound~\cite{vggsound}), where the raw audio is sequentially convoluted by a seises of 1D kernels.
Also, the 1D audio signal can also be transformed to a 2D spectrogram, thus we can adopt the existing popular backbones (\emph{e.g.}, VggNet~\cite{vgg} or ResNet~\cite{res}) instead, where the audio signal's 2D spectrogram can be visualized in the middle of Fig.~\ref{fig:AudioPro}.
To make the audio's 2D spectrogram more discriminative and sensitive to our human auditory system, we can use the Mel filter --- a predefined linear transformation~\cite{mel}, to convert the 2D spectrogram to Mel spectrogram (see the right part of Fig.~\ref{fig:AudioPro}).

\subsection{\textbf{A}udio-\textbf{v}isual \textbf{C}orrespondence (AVC)}
\label{sec:AVC}
As a young task proposed since 2017~\cite{AVCLLL-ICCV17}, the AVC task takes both audio and visual information as input, aiming at making binary predictions on whether the given audio event is synchronized with the current visual event.
For example, a barking dog might be out of the visual field, making the audio event unsynchronized with the visual event. In this case, the AVC task should make a negative prediction, and vice versa.
For a better understanding, we have provided a pictorial demonstration of the AVC task's overview in the bottom-left of Fig.~\ref{fig:Structure}.
Also, Fig.~\ref{fig:AVCpipe} (a) demonstrates the AVC task more clearly from the network perspective.
Clearly, the nature of the AVC task is a typical binary classification, and the technical key is how to align and fuse audio and visual streams.

Arandjelovic~\emph{et al.} in~\cite{AVCLLL-ICCV17} followed an identical network structure to that of Fig.~\ref{fig:AVCpipe} (a). Instead of focusing on the feature representation aspect, the primary interest of this work is to learn the relationship between single static frames and their audio counterparts. To fuse deep features respectively derived from audio and visual sources, the authors have resorted to a series of feature reshape layers (\emph{i.e.}, pooling layers). Hence, both features of audio and visual streams are reshaped to an identical size, which will be later fused via multiple fully-connected layers. The proposed training process requires no additional supervision data, where image and audio training instances pairs are automatically obtained by sampling two different videos, \emph{i.e.}, picking a random frame from video-1 and a random 1-second audio clip from video-2, and please see Fig.~\ref{fig:AVCpipe} (b) for more details. Note that, as the default training protocol, this strategy has been widely used in our AVC research community.

Following the bi-stream structure also, the same authors in~\cite{OTS-ECCV18} have made one significant modification regarding the audio-visual fusion part. In the early version~\cite{AVCLLL-ICCV17}, features respectively derived from audio and visual streams are simply fused via the widely-used feature concatenation operation. However, the concatenation-based fusion tends to misalign both audio and visual signals, resulting in the fused audio-visual features being inadequate for cross-modal retrieval. Thus, \cite{OTS-ECCV18} has adopted the Euclidean distance-based fusion scheme to enforce the feature alignment process.

%\begin{figure}[!h]
%\centering
%\includegraphics[width=0.8\linewidth]{Coatt.png}
%\caption{Co-attention-based audio and visual feature fusion, where the outputs of the co-attention operation can be regarded as the upgraded versions, \emph{i.e.}, $\rm A+$ means upgraded audio features and $\rm V+$ denotes upgraded visual features, where all those clearly not corresponding information can be effectively excluded.}
%\vspace{-0.2em}
%\label{fig:coattfusion}
%\end{figure}

Also using the bi-stream framework, Cheng~\emph{et al.} in~\cite{AVCTSF-MM20} have presented a fancy fusion scheme, where deep features respectively derived from either the audio steam or the visual stream are firstly combined by the newly designed ``co-attention'' operation, which has been shown in Fig.~\ref{fig:AVCpipe} (c).
The primary objective of this co-attention operation is two-fold: 1) enhance audio-visual consistencies and 2) suppress those inconsistencies.
As shown in Fig.~\ref{fig:AVCpipe} (c), the outputs of the co-attention operations can be regarded as the upgraded versions of the original input, \emph{i.e.}, $\rm \textbf{A}+$ and $\rm \textbf{V}+$, where all those clearly unsynchronized information can be effectively excluded. In addition, the exact implementation of co-attention could be either the widely-used spatial attention~\cite{Coatt-WWG-CVPR19} or the fancy transformer~\cite{Transformer-NIPS17}.

To further promote SOTA performance, the existing learning strategies (\emph{e.g.}, contrastive learning~\cite{AVCCL3-ARXIV21}) can be used directly. Morgado \emph{et al.} in~\cite{AVCCL1-ARXIV21} have applied contrastive learning to the AVC task, whose core idea can be briefly summarized as increasing the inter-class distance and decreasing the intra-class distance. In the implementation, training instances belonging to the intra-class are audio and visual pairs whose semantical feature distances are below the given hard threshold. And the rest of the audio-visual pairs are the inter-class cases. There also exist some other similar works (\emph{e.g.},~\cite{AVCCL2-ARXIV21}) which have adopted the existing learning strategies targeting better audio-visual feature embedding. W.r.t., the embedding aspect, we will provide a brief review in the next subsection.

\begin{figure}[!t]
\centering
\includegraphics[width=1\linewidth]{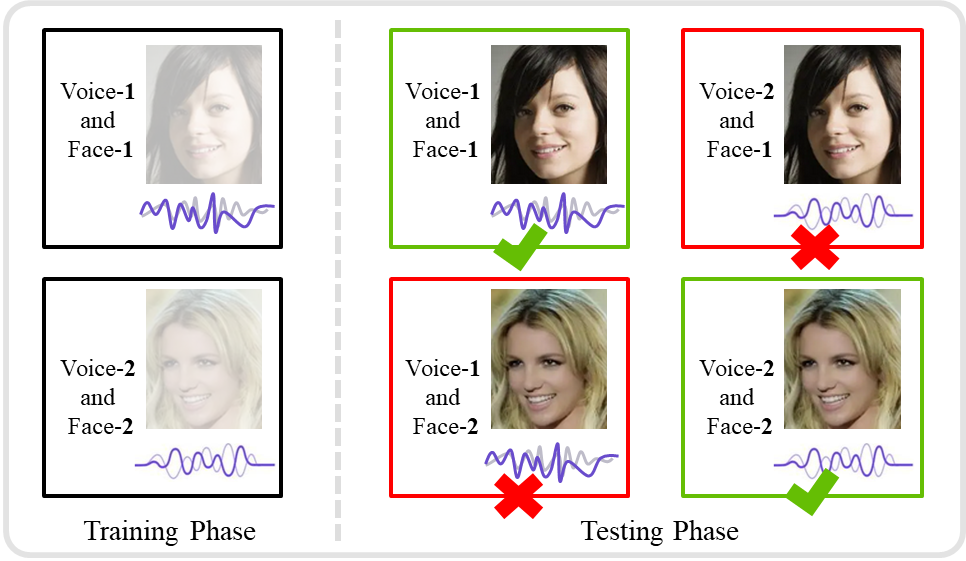}
\caption{Matching between faces and voices. The matched faces and voices pairs are set as positive ({\color{green}\ding{51}}), where the unmatched faces and voices pairs are set as negative ({\color{red}\XSolidBrush}).}
\vspace{-0.2em}
\label{fig:faceandvoice}
\end{figure}

\subsection{Audio-Visual Feature Embedding}
Recently, there have been several works that have focused on the audio-visual feature embedding~\cite{AVCCL3-ARXIV21,AVCRepresent1-ICASSP21,AVCRepresent3-NIPS20}.
The main objective of audio-visual feature embedding is to obtain a generic feature representation, and thus in the spanned feature space, the embedded features can be informative and discriminative enough for specific applications, \emph{e.g.}, the multi-modality image retrieval~\cite{AVCRepresent2-ICCV19}.

Tian~\emph{et al.}~\cite{AVCRepresent4-ECCV20} have applied channel-wise attention to help selectively fuse audio and visual features. The motivation is very straightforward, which is based on an assumption, \emph{i.e.}, that either visual or audio features might benefit the subsequent classification task, and thus the one with a higher feature response should be considered more during the fusion. Following this rationale, the channel-wise attention has been applied to both audio and visual streams at the same time, then the exact modality-wise selection is achieved by performing a softmax.
Note that, this channel-wise attention-based multi-modality selective fusion has also been used in some existing VSOD approaches, \emph{e.g.}, the classic MGA~\cite{VSOD-MGA-ICCV19}.
Recently, Gao \emph{et al.} in~\cite{AVCRepresent5-CVPR20} have adopted a distillation network to compute audio-visual features. A teacher network was initially trained on the visual domain only, where the video tags were used as the classification supervision. Then, a student audio-visual network was trained by taking the predictions from the teacher network as its supervision, and thus the learned intermediate features can achieve automatic alignment between audio and visual and finally obtain a strong audio-visual feature embedding.

\begin{figure}[!t]
\centering
\includegraphics[width=1\linewidth]{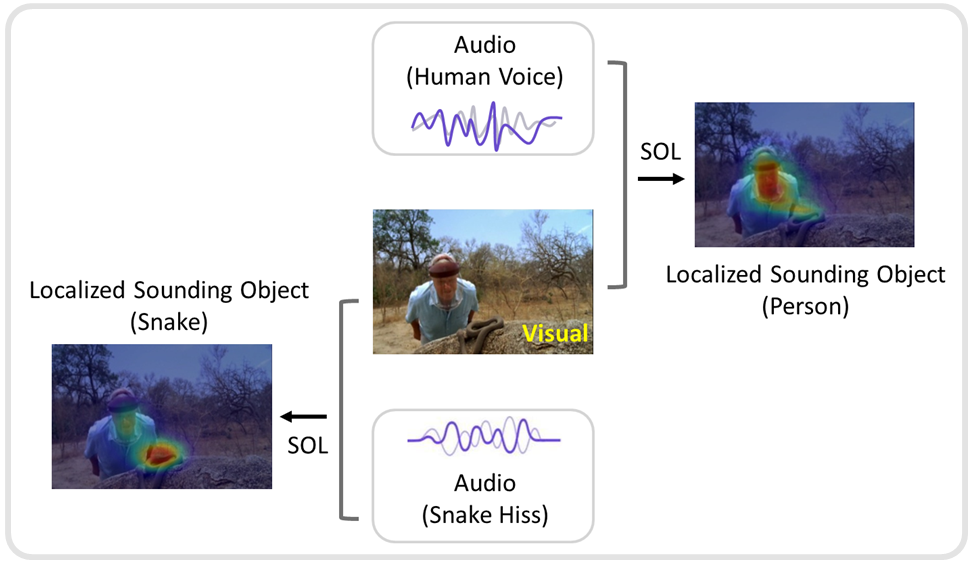}
\caption{The demonstration of the SOL tasks. The objective of SOL is to locate the sounding object in the visual space, \emph{e.g.}, given a visual scene, the snake can be located with the snake hiss, while the person can be located with the human voice.}
\vspace{-0.2em}
\label{fig:SOL}
\end{figure}

The AVC task can also be extended to tell if the current visual information is appropriate with the corresponding audio information.
For example, it is inappropriate for a video frame to contain happy faces yet with a sad melody.
To achieve this goal, Verma \emph{et al.} in~\cite{AVCeMotion-ICASSP19} have ``weakly'' divided the input audio-visual signals into three categories according to their intrinsic emotions, \emph{i.e.}, positive, neutral, and negative, whose structure is almost identical to that of Fig.~\ref{fig:AVCpipe} (c).
A similar solution can be found in~\cite{AVCtheme-ARXIV20}, where the authors have adopted the \emph{video theme} as an additional information source to boost the AVC performance.
The ``video theme'' adopted in this paper is the manual video-level category tags. And the rationale of this work is to use the theme-based classification responses to eliminate instances whose audio-visual semantics are unsynchronized.

Some other representative applications include the \textbf{f}ace and \textbf{a}udio \textbf{m}atching (FAM), and this task's overview can be seen in Fig.~\ref{fig:faceandvoice}. The methodology of the FAM task is quite similar to that of the person \textbf{re}-\textbf{id}entification (ReID)~\cite{Wang-TIP21-Lea,Meng-TPAMI21-Dee,Shu-TCSVT21-Lar,Cao-TMM21-Pro}, while the major difference relies only on their feature modalities, where the ReID task only needs to consider the visual domain, while the FAM task needs to consider both audio and visual.
Also, the key to succeeding in matching faces and voices heavily relies on the design of an appropriate audio-visual fusion.

Similar to the AVC task, Nagrani \emph{et al.} in~\cite{FVM-Conca-CVPR18} directly treated the FAM task as a binary classification. In this work, the face and voice features are obtained by respectively feeding the given face and voice to the existing feature backbone. The audio-visual fusion is simply implemented by the widely-used concatenation operation, and the final classification is fulfilled by the conventional fully-connected layers.
Like the AVC task, the existing learning strategies could also be directly applied to the FAM task and bring solid performance gain, \emph{e.g.}, triplet loss~\cite{FVM-Triplet-ACCV18,FVM-Triplet2-ICMEW19} or contrastive loss~\cite{FVM-Contrast-ECCV18}.
Meanwhile, the FAM research field~\cite{FVM-Embed3-ARXIV17,FVM-Embed2-ECCVW18} has also focused on feature embedding.
Rather than performing the binary classification towards the matching problem, some other weakly-supervised classifications (\emph{e.g.}, identity, gender, and nationality~\cite{FVM-Embed-ECCV18}) towards a single modality can also be used to implicitly obtain the aligned face-voice deep features.

\subsection{\textbf{S}ounding \textbf{O}bject \textbf{L}ocalization (SOL)}
\textbf{Correlation Analysis-based SOL Approaches}.
The objective of SOL is to locate the sounding object in visual space, and the task overview can be seen in Fig.~\ref{fig:SOL}. The research of SOL has a long history, where the earliest work originates in 1999~\cite{SOL-NIPS1999}. In this work, Hershey \emph{et al.}~\cite{SOL-NIPS1999} have explored the correlation between audio and video signals. The idea itself is straightforward, whose rationale is that a spatial region containing a sounding object should have a large probability of exhibiting a strong correlation with the audio signal.
After that, several works have adopted various correlation analysis methods for the SOL task, and we shall briefly review them respectively.

Izadinia \emph{et al.} in~\cite{SOL-MM-TMM13} have applied the \textbf{c}anonical \textbf{c}orrelation \textbf{a}nalysis (CCA)~\cite{SOL-CCA-Bio1936} for identifying the moving objects which are heavily correlated with the audio signal. And similar attempts can be found in~\cite{SOL-OTS-CVPR05,AVE-Dataset-ECCV18}.
Besides, several existing works~\cite{SOL-VLNSS-ACMMM09,SOL-ASIC-PSIVT13} have considered mutual information as the alternation, whose rationales are very similar to that of the CCA-based ones.
In a word, the correlation analysis-based approaches are usually hand-crafted ones, which can only perform well when visual information is really having strong consistency with the audio counterpart.

Different from the correlation analysis-based approaches, which are mainly interested in the SOL task and designed mainly for videos with plain audio signals, there also exist several works~\cite{SOL-Phy1-ICRA08,SOL-Phy2-ICIRS11} which have investigated the stereo cases, \emph{i.e.}, videos with a stereo audio signal.
The key idea of this branch of work is very simple --- the sounding object's spatial location can be coarsely determined by analyzing the difference between the individual soundtracks. Let's take the dual soundtrack for example, the audio signal of a sounding object should first arrive at the left microphone if the sounding object is located on the left.
Theoretically, this type of approaches can achieve the best SOL performance. However, the requirement of the stereo audio signal inevitably narrows the broad applications.

\begin{figure}[!t]
\centering
\includegraphics[width=1\linewidth]{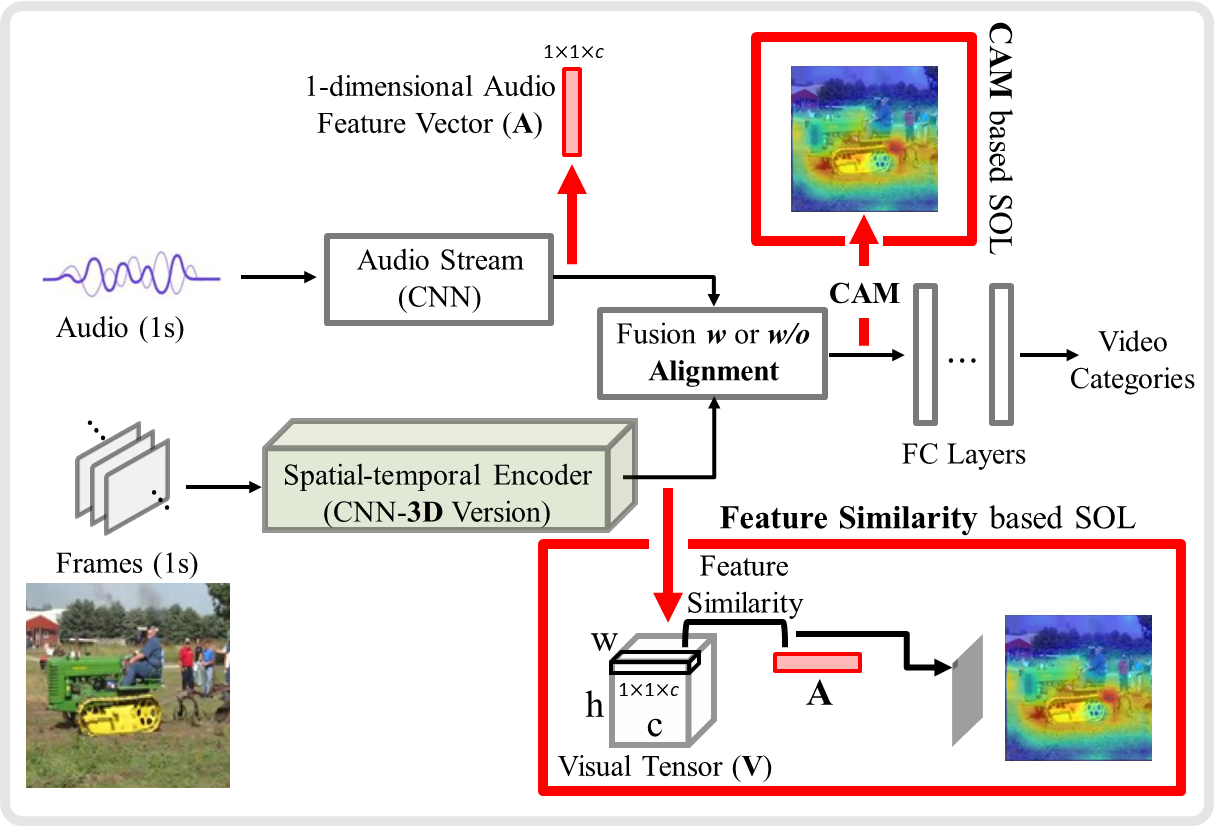}
\caption{Demonstration of the CAM-based SOL and the feature similarity-based SOL. The former searches for strong audio-visual feature response to localize the sounding object, while the latter uses the pixel-wise audio-visual feature similarity instead.}
\vspace{-0.2em}
\label{fig:SOL2}
\end{figure}

\vspace{0.2cm}
\textbf{Deep Learning-based SOL Approaches}.
Recently, SOL-related works are all based on deep learning~\cite{Zhang-IROS21-AcousticFusion,Hu-TPAMI21-CA,Chen-PR21-MF,Chen-CVPR21-LVS}, whose key idea is to perform audio and visual feature embedding. And most of them can be roughly divided into two groups: 1) the \textbf{c}lass \textbf{a}ctivation \textbf{m}apping- (CAM-) based ones and 2) the feature similarity-based ones.

The CAM-based approaches~\cite{AVE-Dataset-ECCV18,SOL-CAM1-CVPR18,SOL-CAM2-TPAMI21,SOL-CAM3-ECCV18} usually adopt the conventional classification network, \emph{e.g.}, the image scene classification. Outwardly, the primary objective of their network training is to achieve a good classification performance. Actually, the real purpose is to utilize the classification task to formulate the audio-visual feature embedding.
Because the sounding objects' audio signal can significantly contribute to the classification task, we can infer that image regions with strong audio-visual feature responses tend to comprise the sounding object.
Following this rationale, the CAM-based SOL methods can directly utilize the feature response map provided at the fusion module's last layer as the SOL result.
Since the CAM's computation is fully automatic, the nature of the CAM-based SOL methods is definitely implicit.

The feature similarity-based SOL methods~\cite{AVCLLL-ICCV17,OTS-ECCV18,SOL-FS1-ICCV20} are slightly different from the CAM-based ones.
Instead of using the implicit manner, this branch of work has adopted the explicit way.
That is, after the classifier training, two separate deep feature representations can be derived from the feature backbones' (\emph{e.g.}, Vgg and VggSound) bottom layers, \emph{i.e.}, a deep visual feature (a 3-dimensional tensor) and a deep audio feature (a 1-dimensional vector).
Then, because those pixels belonging to the sounding object tend to have a strong audio-visual correlation, the pixel-wise audio-visual feature similarity (\emph{e.g.}, the widely-used Euclidean distance and Cosine similarity) can be applied to locate the sounding pixels.
To facilitate a better understanding, we have provided a pictorial demonstration in Fig.~\ref{fig:SOL2}.

\section{Audio Related Saliency Detection}
\subsection{\textbf{A}udio \textbf{S}aliency \textbf{D}etection (ASD)}
Different from the 2-dimensional visual signal, the audio signals can only carry temporal information representing signals' variations over the time scale.
Thus, it is very difficult to build a clear spatial alignment from audio to visual.
By considering audio solely, saliency detection can still be performed, \emph{a.k.a.}, audio saliency detection or salient event detection, and there exist multiple works~\cite{Nakatani-ACCESS22-AS,Wang-TCDS22-Fun}, most of which are non-deep learning-based ones, and we shall briefly review them here.

Following the rationale proposed in the earliest Itti's classic work~\cite{Itti-VR2000} --- salient regions should exhibit high contrast to their surroundings, Kayser~\emph{et al.} in~\cite{AS-CB2005} have investigated the audio saliency detection task.
In this work, the authors have adopted multiple filters to measure the audio signal's changing tendency, \emph{i.e.}, the first derivative of intensity and frequency over the time scale.
Because, for a short time span, salient audio fragments usually come with a large difference from the rest, their temporal-scale changing tendency can be very effective in evaluating saliency.
Following a similar idea, Schauerte~\emph{et al.}~\cite{AS-ICASSP13} have adopted the KL-divergence between two audio fragments' 2D spectral histograms. Compared with the previous work~\cite{AS-CB2005} which could be regarded as a ``local'' audio saliency approach, this new work is clearly a non-local one.
Also, based on the 2D spectral histogram, Tsuchida~\emph{et al.} in~\cite{AS-PAMCSS12} have proposed a novel non-local signal feature representation method. For each cell in 2D spectral histogram, the authors have used \textbf{p}rincipal \textbf{c}omponent \textbf{a}nalysis (PCA) to extract the non-local feature. Based on these features, audio saliency can be obtained by performing contrast computation over the newly devised feature subspace.

Also, the audio signal's amplitude and frequency are the widely-used computational unit for the salient event detection~\cite{Rodriguez-ACCESS18-The}. Zlatintsi~\emph{et al.} in \cite{AS-ESPC12} have converted both audio amplitude and frequency to 3D feature via the Teager energy~\cite{AS-ICASSP09}. This work has made a very strong assumption that people tend to be attracted by sudden loudness.
Thus, the authors have directly considered the averaged audio's amplitude, frequency, and newly devised energy to measure the saliency degree.
Beyond the amplitude and frequency-based representations, Merve~\emph{et al.} in~\cite{AS-CISS12} have further devised several novel feature representations (\emph{e.g.}, envelope feature, bandwidth feature, rate feature, pitch feature) for audio signal over the time scale. Finally, this work follows the conventional common thread, \emph{i.e.}, the contrast computation, for each of the newly devised features to obtain multiple bottom audio saliency, and the final saliency is achieved by combining all of them via simple linear fusion.

There also exists a research branch focusing on the relationship between text information and audio saliency~\cite{AS-ICASSP12,AVS-TMM13}.
Different from the above-mentioned audio saliency methods which consider audio signal only, this branch has additionally considered the text information.
Zlatintsi~\emph{et al.} in~\cite{AS-EUSIPCO15} have fused both text information and audio signal before computing the audio saliency, where the key rationale of the adopted fusion is to compute the feature similarity (\emph{e.g.}, mutual information) between text and audio.
Another most representative work could be the~\cite{AS-ICASSP19}, which has adopted a non-negative matrix factorization model to measure the consistency between text and audio.

In a word, the advances toward audio saliency detection are relatively slow, where the widely-used methodologies are still limited to the conventional hand-crafted ones, and the deep learning-related researches are quite rare.
Considering the importance of audio saliency, this field really deserves intense research attention in the near future.

\subsection{\textbf{A}udio-\textbf{v}isual \textbf{S}aliency \textbf{D}etection (AVSD)}
As the main topic of this review, we shall give a more detailed introduction and discussion of the SOTA audio-visual saliency detection approaches.
However, to the best of our knowledge, this topic is definitely in its infancy, and only several deep learning-based works exist. Thus, we decide to take the exact audio-visual fusion scheme as the starting point. Thus, some related works mentioned in the previous sections might be referenced here for a better understanding.

\vspace{0.2cm}
\textbf{Hand-crafted Naive Fusions for AVSD}.
Most of the existing hand-crafted approaches~\cite{AVS-CISS13,AVS-ISEI16} follow the bi-stream structure, which is almost the same as the AVC task reviewed above. Given a video sequence, saliency detection over the either audio or visual channel is computed first, then the audio-visual saliency can be derived by designing an appropriate fusion scheme.
Clearly, any off-the-shelf audio/visual saliency detection methods can be used directly, making the exact fusion scheme the key to the overall performance.

Many works~\cite{Zhu-ICME21-Lavs,Yao-ICIP21-DAV,Chao-VCIP20-Tow,Akolkar-EBCCSP15-EDVS} have simply adopted the multiplicative-based fusion, because it can effectively enhance the consistency and compress the inconsistency between audio and visual saliency-related features, because those real salient regions tend to be salient in both audio domain and visual domain simultaneously.
The limitation of the multiplicative-based fusion is also quite clear --- it tends to get confused if there exist multiple visual and saliency features.
In cases with multiple audio and visual features, Coutrot~\emph{et al.} in~\cite{AVS-ICIP14} have adopted the linear fusion, where the fusion weights are computed via the classic \textbf{e}xpectation-\textbf{m}aximization (EM) algorithm, a statistical method using training samples to estimate the relative importance of each feature aiming to maximize the global likelihood of the mixture model.
Further, Sidaty~\emph{et al.} in~\cite{AVS-NC17} have conducted an extensive evaluation regarding different fusion schemes, including maximum, addition, average, multiplication, and non-linear combination-based fusion schemes.
As expected, all such simple fusions are inferior to the non-linear fusion, because, in most cases, the audio saliency and the visual saliency could have different contributions to the final audio-visual saliency, and the exact contribution degree is usually determined by the given video scene and content, yet these naive fusion schemes are not flexible enough, failing to achieve the optimal balance between audio and visual.

\begin{figure*}[!t]
	\centering
	\includegraphics[width=0.75\linewidth]{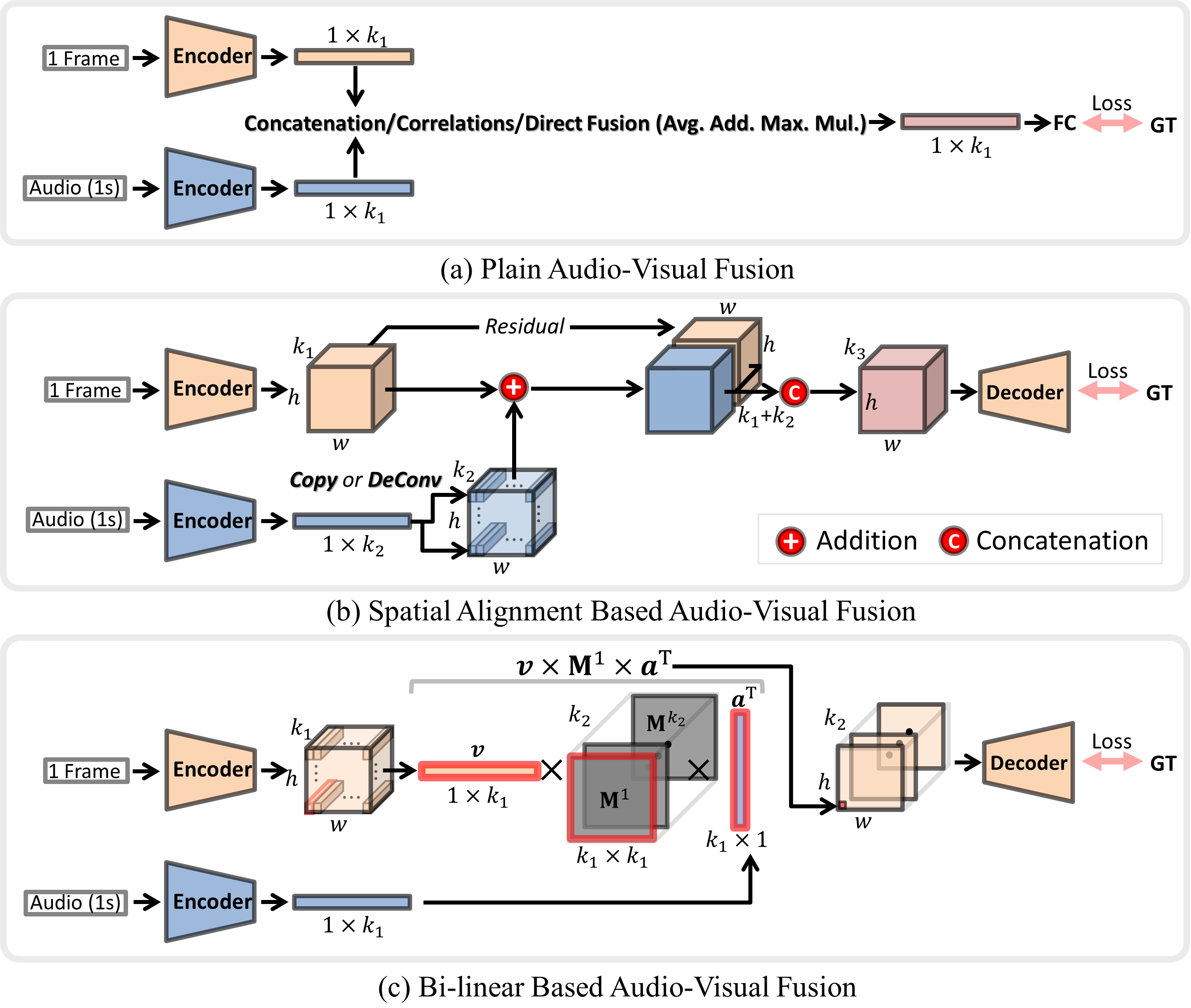}
	\caption{The most representation fusion schemes for audio-visual saliency detection. Among them, (a) merely utilizes the conventional plain concatenation operations to integrate audio and visual features; (b) treats the audio part as auxiliary information, and the embedded semantical consistency is used to highlight the corresponding spatial regions; (c) adopts a dimension transformation matrix to handle the dimension mismatched problem, which doesn't require the identical dimension size of the individual audio and visual saliency cues.}
	\vspace{-0.2em}
	\label{fig:FusionAll}
\end{figure*}

\vspace{0.2cm}
\textbf{Correlation Analysis-based Fusions for AVSD}.
From the experiment perspective, Min~\emph{et al.} in~\cite{AVS-QOME14} have conducted extensive verifications of the human eye fixations in conditions with and without audio signals. Their results indicate that audio signals can significantly influence human attention only if the salient object is visually non-salient yet salient in the audio channel; otherwise, the audio information is completely helpless.
This work also inspires us that an audio-visual saliency detection method should bias more towards visual signals in most cases.
Following the same rationale, Min~\emph{et al.} in~\cite{AVS-TOMM16} have adopted the classic \textbf{c}anonical \textbf{c}orrelation \textbf{a}nalysis (CCA) to localize spatial regions which have demonstrated strong audio-visual consistency.
Since the audio and visual saliency cues have been computed in advance, the fusion process mainly targets at highlighting the visual regions correlated well to the audio.
More recently, Min~\emph{et al.} in~\cite{AVS-TIP20} have further considered the deep learning-based saliency cues. And the CCA has been replaced by its upgraded variant --- the \textbf{k}ernel \textbf{c}anonical \textbf{c}orrelation \textbf{a}nalysis (KCCA), to measure the audio-visual correlation.
The main reason is that the CCA can only correlate linear relationships, while the KCCA is able to map features to higher-dimensional feature spaces and increase the nonlinearity, which could be more practical in the audio-visual saliency detection task.

\vspace{0.2cm}
\textbf{A Brief Summary of Hand-crafted Fusions}.
To further explore the advantages and disadvantages of the existing fusion schemes, Tsiami~\emph{et al.} in~\cite{AVS-SPIC19} have compared three widely-used audio-visual fusion schemes, \emph{e.g.}, direct fusion (\emph{i.e.}, the multiplicative-based fusion), linear correlation coefficient~\cite{AVS-MR13}, and mutual information~\cite{AVS-TAMD09}.
The authors have combined the existing visual saliency models with the off-the-shelf audio saliency models by using one of these fusion schemes alternatively, and the quantitative results have reached a clear conclusion, \emph{i.e.}, the exact optimal fusion scheme is determined by multiple factors, including both the quality of low-level saliency cues and the input video data.
For ``raw'' hand-crafted saliency cues computed by models which are good at measuring saliency from the temporal scale, the correlation coefficient could be the best choice, since it mainly considers the temporal consistency between audio and visual.
As for the case where the raw saliency cues have been incorporated with spatial information, mutual information could be the optimal choice.
However, things could be changed for those ``refined'' saliency cues --- saliency cues obtained via deep learning-based top-down models, where the direct fusion usually exhibits the best fusion performance, because the refined saliency cues are usually more trustworthy than those raw ones, and thus they could be directly used to complement their counterparts.

\vspace{0.2cm}
\textbf{SOTA Deep Learning-based AVSD Methods}.
After entering the deep learning era, massive deep learning-based visual saliency models have been proposed. However, to the best of our knowledge, there only exist five deep learning-based audio-visual saliency detection models~\cite{AVS-DAVE20,AVS-STAVIS20,AVS-AVI21,AVS-WGT21,Chen-NC21-ASDL}. Here we shall provide a detailed review regarding these works respectively.
For a better understanding, we have provided multiple method pipelines to clarify the audio-visual fusion methodology regarding these SOTA deep learning-based audio-visual saliency detection works.
As can be seen in Fig.~\ref{fig:FusionAll}, all three sub-figures respectively correlate to the SOTA models mentioned above: sub-figure (a)~\cite{AVS-DAVE20,Chen-NC21-ASDL}, sub-figure (b)~\cite{AVS-WGT21}, and sub-figure (c)~\cite{AVS-AVI21,AVS-STAVIS20}.

We shall first introduce the~\cite{AVS-DAVE20,Chen-NC21-ASDL}. As shown in Fig.~\ref{fig:FusionAll} (a), the audio-visual fusion adopts the conventional plain concatenation operations, which takes both audio and visual feature tensors as input, and the saliency predictions are obtained via a typical decoder after concatenating both audio and visual tensors. Specifically, because the audio modality has a completely different formation from the visual modality, it is required to ensure that the audio's tensor feature has the same size as its visual counterpart. Clearly, the overall method rationale of this work is very straightforward, and the concatenation-based fusion could be replaced by other existing ones, \emph{e.g.}, direct fusions and correlation analysis tools, where similar works have been widely adopted by the AVC task which have been reviewed in Sec.~\ref{sec:AVC}.

As illustrated in Fig.~\ref{fig:FusionAll} (b), the spatial alignment-based audio-visual fusion can bias the fusion toward the visual part, where the deep audio feature, which usually is a 1-dimensional vector with the same size as the visual tensor's channel number, is either de-convolved or copied to correlate to each spatial location.
This implementation has treated the audio as auxiliary information, where the embedded semantical consistency is the key factor in highlighting the corresponding spatial regions as the salient ones.
The ``copy'' scheme has also been widely used by the AVC task. However, to the best of our knowledge, \cite{AVS-WGT21} is the first attempt to use de-convolution for the audio-visual alignment.
Also, either the copy or the de-convolution-based alignment can be combined with the popular ``residual'' operation to focus the fusion process on the visual signal, because, in most cases, the visual signal is definitely stronger in determining human attention than the audio signal.

Lastly, as demonstrated in Fig.~\ref{fig:FusionAll} (c), we introduce the bi-linear audio-visual fusion, which has been adopted by~\cite{AVS-STAVIS20,AVS-AVI21} and achieved the leading SOTA performance. Compared with either the plain or spatial alignment-based fusion, the bi-linear fusion has one significant advantage: it doesn't require the individual audio and visual saliency cues to have an identical dimension size, where a dimension transformation matrix, \emph{i.e.}, see the $\textbf{M}$ in the sub-figure C, is adopted to handle the dimension mismatched problem.
Actually, the bi-linear fusion also has its own limitation, \emph{i.e.}, the semantical correspondence between audio and visual channels has been destroyed, making the modeling of complex audio and visual interactivity very difficult.
In sharp contrast, the spatial alignment-based fusion could make full use of the semantical information provided by either off-the-shelf visual (\emph{e.g.}, ResNet50) or audio (\emph{e.g.}, VggSound) feature backbone, where the learned semantical information could shrink the problem domain effectively. As a result, the audio-visual complementary fusion status could be easily reached even in a complex audio-visual environment.

\begin{figure}[!t]
	\centering
	\includegraphics[width=1\linewidth]{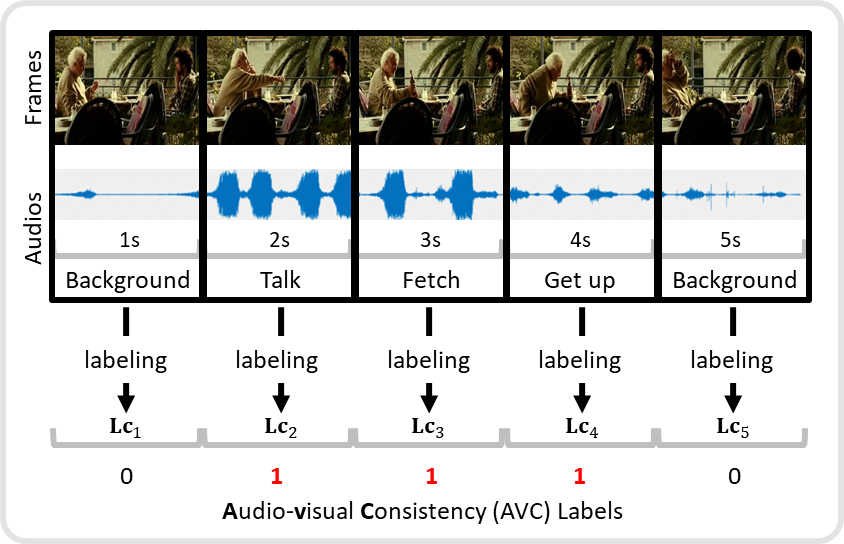}
	\caption{Detailed demonstration of our AVC annotation. In the selected video clip, when two men are talking, the sounds of 1st and 5th seconds are the background music, whose AVC labels are set to 0, meaning that the audio and the visual are semantically mismatched. While, from the 2nd to the 4th seconds, the audio signals are talking sound, fetching sound, and getting up sound respectively, and thus we label them as 1 because these audio-visual fragments are clearly matched.}
	\vspace{-0.2em}
	\label{fig:annotation}
\end{figure}

\section{Audio-visual Semantical Consistency Perceptual}

\subsection{Preliminary}
Existing \textbf{a}udio-\textbf{v}isual \textbf{s}aliency \textbf{d}etection (AVSD) works mainly adopt bi-stream network architecture, where audio saliency and visual saliency are computed individually and combined later as the final output.
In fact, when the audio signal is not consistent with the visual signal, the audio saliency is completely helpless to complement the visual saliency, which takes up about 60\% of all cases.
For example, in an image, two persons are talking while the background music comes from the outside, and, in this case, the audio signal cannot benefit the visual in determining saliency.

Inspired by previous multimedia related works~\cite{Chen-TASLP21-TSC,Chang-TAFFC21-esc,Han-ICASSP20-CDCU}, we propose to introduce the ``\textbf{a}udio-\textbf{v}isual \textbf{c}onsistency (AVC)'' into our saliency detection research field.
The major highlight of our approach is its generic usage, which is capable of upgrading any SOTA bi-stream-based AVSD model from ``AVC-unaware'' to ``AVC-aware''.
In fact, the optimal audio-visual fusion is very difficult to achieve if the adopted AVSD model is AVC-unaware, because the model is completely blind and thus cannot completely omit the audio when the audio is not corresponding to the visual.
Thus, in facing helpless audio signals, an AVSD model taking both audio and visual is clearly inferior to the model using the visual solely, yet this ``binary switch'' cannot be achieved if the model is AVC-unaware.

An intuitive way to convert an AVC-unaware AVSD model to AVC-aware is to resort to an additional module which can automatically predict whether the currently given audio is consistent with the visual.
Therefore, to fully realize our idea --- making any existing bi-stream AVSD model AVC-aware, two things should be prepared in advance: 1) train the aforementioned classifier, and 2) integrate the classifier into the AVSD model.
Next, we shall respectively detail each of them in the following subsections.

\begin{figure*}[!t]
	\centering
	\includegraphics[width=0.8\linewidth]{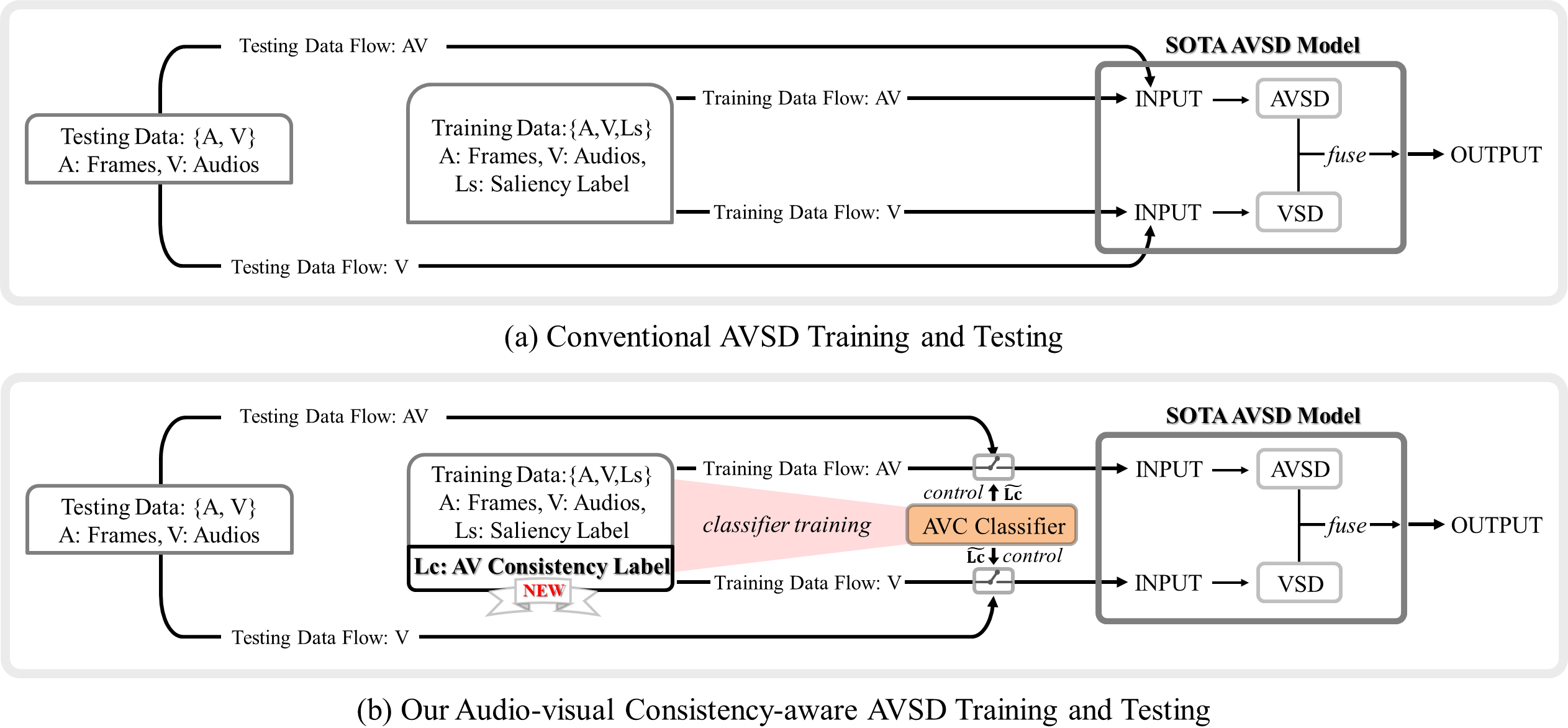}
	\caption{Demonstrations of the differences between the conventional audio-visual saliency detection model training/testing pipeline (a) and the newly modified training/testing pipeline (b). The advocated AVC classifier can be trained by the newly annotated AVC labels, and then dynamically control the data flow of the adopted bi-stream SOTA AVSD model. The $\tilde{\textbf{Lc}}$ denotes the binary output of the AVC classifier. Clearly, by equipping the existing SOTA AVSD model with the AVC classifier, we can make the original AVC-unaware AVSD model AVC-aware, achieving persistent performance improvement.}
	\vspace{-0.2em}
	\label{fig:pipeline}
\end{figure*}

\subsection{Audio-visual Consistency Labeling}
The AVC classifier can of course be trained via the above-mentioned weakly-supervised method, which has been shown in Fig.~\ref{fig:AVCpipe} (b).
However, the overall performance of this method is usually too limited to benefit the saliency detection task.
Thus, we propose to utilize the fully-supervised method to train the audio-visual consistency classifier.

To achieve this goal, we shall manually equip each video frame with AVC labels.
That is, we manually provide all the existing benchmark AVSD datasets with binary AVC labels, and a representative pictorial demonstration has been shown in Fig.~\ref{fig:annotation}. Thus, each audio-visual fragment will be assigned to 1 or 0 label accordingly.
Suppose all existing training instances (with $N$ frames) can be represented as: $\rm \{{A}_i,{V}_i,{L}s_i\}$, where $i\le N$, A and V respectively denote the audio and visual, and $\rm {L}s$ is the corresponding fixation map.
During the annotation process, if the audio sound is made by the salient object\footnote{We manually regard an object as salient if it has the highest fixation number in the scene.}, we regard that the audio and visual are consistent, thus we assign the AVC label as 1. Otherwise, if the audio is unseen background music or off-screen sound, the AVC label of this audio-visual fragment is set as 0.
For a better understanding, we have provided a pictorial demonstration regarding how to perform the proposed AVC labeling process, which can be found in Fig.~\ref{fig:annotation}.

After the annotation process\footnote{To match the fps of video clips (25$\sim$30) and the audio length, we resort to Adobe Premiere CC, a professional video editing software, to align the mismatched audio and visual durations.}, each training instance can be converted to:
\begin{equation}
	\rm \{A_i,V_i,Ls_i\}\rightarrow \{A_i,V_i,Ls_i,\textbf{Lc}_i\},\ \ \ \textbf{Lc}_i\in\{0,1\}, i\in[1,\emph{N}],
	\label{eq:score}
	\vspace{-0.1cm}
\end{equation}
where $\textbf{Lc}$ denotes the newly annotated audio-visual consistency label, and $N$ denotes the total frame number.
We have newly annotated all publicly available AVSD benchmarks, totally 5 sets (or 6 if Coutrot set is divided into Coutrot1 set and Coutrot2 set) consisting of 241 video clips involving 300,000 frames.
These newly annotated datasets are now publicly available\footnote{\url{https://github.com/songsook/SCDL}}.

%Notice that, in our method, the newly provided AVC labels can be used to automatically filter out those inconsistent audio and video pairs.
%Thus, only those consistent pairs are used for the bi-stream joint training, and
%Only consistent pairs are chosen to train the full AVSD models, while those inconsistent pairs are fed into the visual stream, which can promote the AVSD models' performance enormously.

\begin{table}[!t]
  \caption{Details of the existing AVSD sets.\vspace{-0.2cm}}
	\begin{center}
	\renewcommand{\arraystretch}{1.1}{
		\setlength{\tabcolsep}{5.5pt}{
			\resizebox{0.9\linewidth}{!}{
    \begin{tabular}{r|c|c|c|r|l}
    \Xhline{1.0pt}
    \hline
  %  \specialrule{1.0pt}{0pt}{0pt}
    Datasets & Year & Videos & Viewers & Frames & Links\\
     \hline
     \hline
    DIEM~\cite{diem} & 2010 & 84 & 42 & 78,167 & [\href{https://thediemproject.wordpress.com/}{Link}]\\
    AVAD~\cite{avad} & 2016 & 45 & 16 & 9,564 & [\href{https://sites.google.com/site/minxiongkuo/home}{Link}]\\
    Coutrot1~\cite{coutrot1} & 2013 & 60 & 72 & 25,223 & [\href{http://www.gipsa-lab.fr/}{Link}]\\
    Coutrot2~\cite{coutrot2} & 2014 & 15 & 40 & 17,134 & [\href{http://antoinecoutrot.magix.net/public/databases.html}{Link}]\\
    SumMe~\cite{summe} & 2019 & 25 & 10 & 109,788 & [\href{https://gyglim.github.io/me/vsum/index.html}{Link}]\\
    ETMD~\cite{etmd} & 2019 & 12 & 10 & 52,744 & [\href{http://cvsp.cs.ntua.gr/research/aveyetracking/}{Link}]\\
    \hline
  %  \specialrule{1.0pt}{0pt}{0pt}
    \Xhline{1.0pt}
	\end{tabular}}}}
  \label{tab:Datasets}%
  \vspace{-0.4cm}
	\end{center}
\end{table}%

\subsection{The Proposed AVC-aware AVSD Model}
The conventional \textbf{a}udio-\textbf{v}isual \textbf{s}aliency \textbf{d}etection (AVSD) training and testing protocol has been shown in Fig.~\ref{fig:pipeline} (a), where the AVSD model is a typical bi-stream fusion net, which combines its AVSD and \textbf{v}isual \textbf{s}aliency \textbf{d}etection (VSD) to formulate the final result.
Clearly, the VSD stream is the mainstream, and the AVSD stream is the auxiliary stream, where audio and visual are fused early via fusion schemes mentioned in Fig.~\ref{fig:FusionAll}, to further promote the VSD stream.
As we have mentioned before, this typical AVSD training and testing protocol is completely AVC-unaware, the later fusion (\emph{i.e.}, fuse VSD with AVSD) could even degenerate the overall performance when the given audio and visual are mismatched.

\begin{figure*}[!t]
	\centering
	\includegraphics[width=1\linewidth]{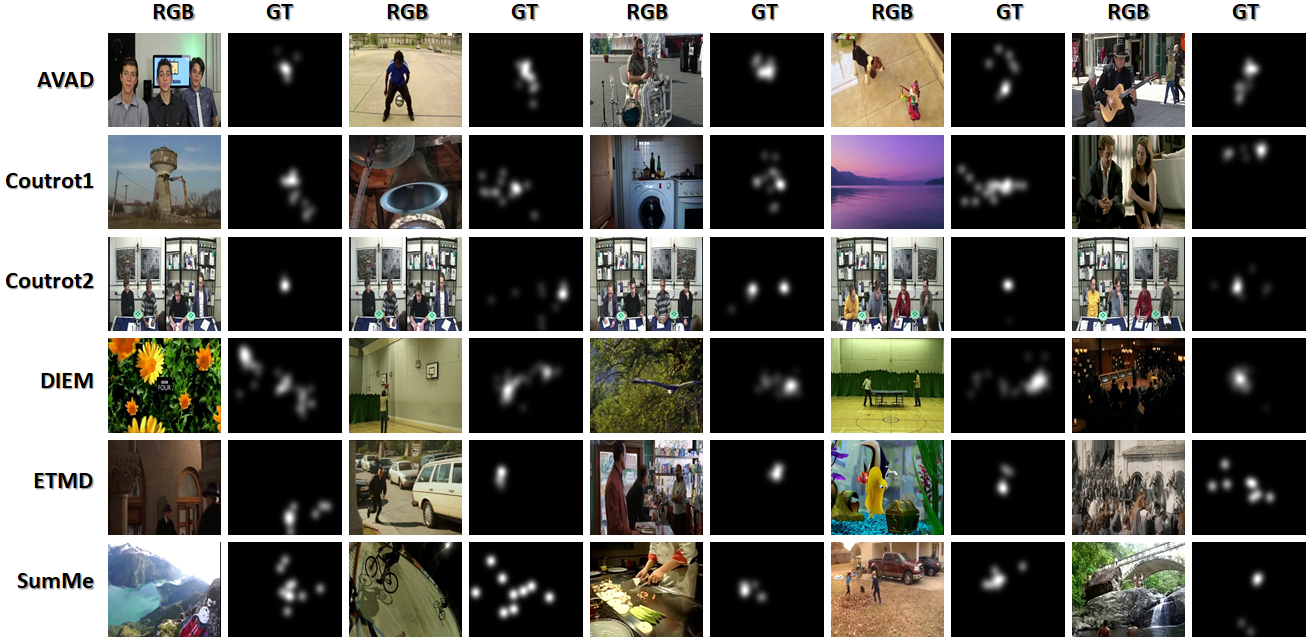}
	\caption{Demonstration of the differences regarding the scene contents of six wide-usedly datasets of AVAD, Coutrot1, Coutrot1, DIEM, ETMD and SumMe.}
	\vspace{-0.2em}
	\label{fig:dataset}
\end{figure*}

To handle the above-mentioned problem, we propose the AVC-aware training and testing protocol, which has been shown in Fig.~\ref{fig:pipeline} (b), whose major difference to (a) is the newly provided AVC classifier, and this classifier can be trained by using the newly equipped AVC labels.
In our implementation, we simply use an identical classifier structure to AVID~\cite{AVCCL1-ARXIV21} to predict the AVC degree of the current input audio-visual fragment automatically, outputting 0 or 1.
Notice that other classifier structures can also be used, and we have tested several others, where the quantitative result (Table~\ref{ablation}) suggests that the AVID is the best choice.

As shown in Fig.~\ref{fig:pipeline} (b), the newly proposed AVSD model can be trained in the typical end-to-end way, where the AVC classifier serves the existing SOTA bi-stream AVSD model, \emph{i.e.}, Fig.~\ref{fig:pipeline} (b), as ``binary switchers'' to control the INPUT of the adopted SOTA AVSD model.
In other words, the output of the AVC classifier determines whether or not the single V flow or both V and AV flows are to be used in the subsequent SOTA AVSD model.
That is, when the output of the AVC classifier is 0, which means the current audio is inconsistent with the current visual, suggesting removing the AV flow from fusing it with V flow, because, for an inconsistent audio-visual fragment, the output of AV flow tends to significantly inferior to the output of V flow, thus fusing AV with V MUST degenerate the overall performance.
When the output of the AVC classifier is 1, the whole training process is completely identical to the original SOTA AVSD model, where both AV flow and V flow are simultaneously considered.
The entire data flow of our AVC-aware AVSD model can be expressed as:
\begin{equation}
\begin{aligned}
{\rm OUTPUT}&\gets \tilde{\textbf{Lc}} \cdot {\rm Fuse}({\rm AV},{\rm V}) + (1-\tilde{\textbf{Lc}})\cdot {\rm V}, \\
	\tilde{\textbf{Lc}} &= \rm{AVC}_{cls}({\rm AV},{\rm V})\in\{0,1\},
\label{eq:score}
\end{aligned}
\end{equation}
where $\rm{AVC}_{cls}$ represents the AVC classifier, and $\tilde{\textbf{Lc}}$ is the binary prediction regarding AVC of the current input V and AV.

The training process of our AVC-aware AVSD model consists of two tasks, \emph{i.e.}, 1) the conventional audio-visual saliency detection task, which takes the saliency labels (Ls) as GT, and 2) the newly added AVC classifier training, which takes the AVC labels (Lc) as GT.
Thus, the overall loss function {${\rm L}_{all}$} can be detailed as:
\begin{equation}
	{\rm L}_{all} = (1-\rho)\cdot{\rm L}_{cls} + \rho\cdot{\rm L}_{avsd},
	\label{eq:score}
	\vspace{-0.1cm}
\end{equation}
where ${\rm L}_{cls}$ is a typical cross-entropy loss targeting at the training of AVC classifier, ${\rm L}_{cls}$ is the \textbf{K}ullback-\textbf{L}eibler (KL) divergence loss, the most widely-used loss function in AVSD model training, and $\rho$ is a balancing factor which we empirically assign it to 0.5.
In the testing phase, the exact data flows are dynamically controlled by the AVC classifier in an identical way to the training phase.

In brief, the major highlight of our approach is its generic design, which can serve any existing bi-stream SOTA AVSD models as the plug-in, and promote their performances persistently.
Though a more fancy network design could bring additional performance gain, we shall leave it to future work to stay the main focus of our topic.

\begin{table*}[!t]
	\caption{Quantitative comparisons between our method with other fully-/weakly-/un-supervised methods on all 6 datasets. The best result is marked in \textbf{bold} font. * means that the target models (\emph{e.g.}, STANet*, STAViS*, and AVINet*) are trained by the whole pipeline in Fig.~\ref{fig:pipeline} with AVC classifier; \# denotes that the target models (\emph{i.e.,}, STANet\#, STAViS\#, and AVINet\#) are trained by removing the AV classifier model, and their OUTPUTs are manually reformulated by using \textbf{Lc} (Eq.~\ref{eq:score}) as the indicator, standing for the ideal cases.}
	\vspace{-0.2cm}
	\begin{center}
	\newcommand{\tabincell}[2]{\begin{tabular}{@{}#1@{}}#2\end{tabular}}
		\renewcommand{\arraystretch}{1.2}{
			\setlength{\tabcolsep}{3.5pt}{
				\resizebox{1.0\linewidth}{!}{
					\begin{tabular}{c|r||ccccc|ccccc|ccccc}
						\Xhline{1.0pt}
						\cline{1-17}
						\multirow{2}{*}{Means}&Datasets
						&\multicolumn{5}{c}{{AVAD~\cite{avad}}}\vline &\multicolumn {5}{c}{{DIEM~\cite{diem}}}\vline&\multicolumn{5}{c}{{SumMe~\cite{summe}}}\\
						\cline{2-17}
						&Methods &AUC-J$\uparrow $ 	& SIM$\uparrow $ &s-AUC$\uparrow $ & CC$\uparrow $& NSS$\uparrow $ 	&AUC-J$\uparrow $ 	& SIM$\uparrow $ &s-AUC$\uparrow $ & CC$\uparrow $& NSS$\uparrow $&AUC-J$\uparrow $ 	& SIM$\uparrow $ &s-AUC$\uparrow $ & CC$\uparrow $& NSS$\uparrow $  \\ 	 	
						\hline
						\hline																
						\multirow{4}{*}{\rotatebox{90}{\tabincell{c}{Un-\\surpervised}}}
						&{ITTI~\cite{ITTI}}
						&0.688 &0.170 &0.533 &0.131 &0.611
						&0.663 &0.217 &0.583 &0.137 &0.555
						&0.666 &0.151 &0.559 &0.097 &0.436
 						\\				
						&{GBVS~\cite{Jonathan-GBVS}}
						&\textbf{0.854} &0.247 &0.572 &\textbf{0.337} &\textbf{1.556}
						&\textbf{0.830} &\textbf{0.318} &0.605 &\textbf{0.356} &\textbf{1.277}
						&\textbf{0.808} &0.221 &0.567 &\textbf{0.272} &\textbf{1.134}
						\\							
						&{SBF~\cite{SBF}}
						&0.833 &\textbf{0.272} &0.576 &0.308 & 1.489
						&0.759 &0.292 &0.608 &0.301 &1.081
						&0.783&\textbf{0.228}&0.590&0.230& 1.023
						\\							
						&{AWS-D~\cite{AWS-D}}
						&0.825 &0.221& \textbf{0.589}&0.304& 1.378
						&0.733&0.250&\textbf{0.612}&0.301&1.128
						&0.747&0.192&\textbf{0.603}&0.186&0.853
						\\		
						
						\cline{1-17}
						\Xhline{1.0pt}
						\multirow{6}{*}{\rotatebox{90}{\tabincell{c}{Weakly-\\surpervised}}}
						&{GradCAM++~\cite{Grad-CAM}}
						& 0.777 &0.273& 0.559&0.255&1.217
						&0.732& 0.216&0.583& 0.271&0.778
						&0.774&0.217& 0.593&0.225&0.924
						\\
						&{WSS~\cite{WSS}}
						& 0.858 &0.292& \textbf{0.592}&0.347& 1.655
						&0.803& 0.333&0.620&0.344&1.293
						&0.812&0.245& 0.589&0.279& 1.098
						\\		
						&{MWS~\cite{MWS}}
						&0.834 &0.272& 0.573&0.309& 1.477
						&0.806& 0.336&0.628&0.350& 1.308
						&0.808&0.237& 0.607&0.258&1.155
						\\					
						\cline{2-17}
						&{STANet~\cite{AVS-WGT21}}
						&0.873 &0.334 &0.580 &0.438 &2.018
						&0.861 &0.391 &0.658 &0.469 &1.716
						&0.854 &0.294 &0.627 &0.368 &1.647
						\\
						&\cellcolor[rgb]{1,0.8,0.8}{STANet*}
						&0.879 &\textbf{0.341}&0.584&0.439&2.068
						&0.891&\textbf{0.392}&0.662&\textbf{0.498}&2.016
						&0.870&0.323&0.631&0.382&1.662
						\\
						&\cellcolor[rgb]{0.7,0.8,1.0}{STANet\#}
						&\textbf{0.881}&\textbf{0.341}&\textbf{0.585}&\textbf{0.442}&\textbf{2.070}&\textbf{0.892}&0.390&\textbf{0.665}&\textbf{0.498}&\textbf{2.019}&\textbf{0.873}&\textbf{0.325}&\textbf{0.632}&\textbf{0.384}&\textbf{1.663}\\	
						\cline{1-17}
						\Xhline{1.0pt}
						\multirow{8}{*}{\rotatebox{90}{\tabincell{c}{Fully-\\surpervised}}}
						&{DeepVS~\cite{Jiang-ECCV18}}
						&0.896 &0.391&0.585&0.528&3.010
						&0.840& 0.392&0.625&0.452&1.860
						&0.842&0.262& 0.612&0.317&1.620
						\\ 	
						&{ACLNet~\cite{ACLNet}}
						&0.905 &0.446&0.560&0.580& 3.170
						&0.869&0.427&0.622&0.522&2.020
						&0.868&0.296&0.609&0.379&1.790
						\\ 		
						\cline{2-17}
						&{STAViS~\cite{AVS-STAVIS20}}
						&0.919 &0.457 &0.593 &0.608 &3.180
						&0.883 &0.482 &0.674 &0.579 &2.260
						&0.888 &0.337 &0.656 &0.422 &2.040 	
						\\ 	
						&\cellcolor[rgb]{1,0.8,0.8}{STAViS*}
						&0.925 &0.460&0.599&0.623&3.252
						&0.896&0.484&0.683&0.582&2.499
						&0.903&0.393&0.634&0.460&2.102\\
						&\cellcolor[rgb]{0.7,0.8,1.0}{STAViS\#}
						&0.927&0.463&0.597&0.622&3.255&0.899&0.485&0.684&0.581&2.497&0.904&0.397&0.635&0.463&2.107\\	
						\cline{2-17}
						&{AViNet~\cite{AVS-AVI21}}
						&0.931 &0.499 &0.663 &\textbf{0.689} &3.740
						&0.901 &0.504 &0.722 &0.637 &2.540
						&0.900 &0.350 &0.697 &0.470 &2.420
						\\ 		
						&\cellcolor[rgb]{1,0.8,0.8}{AViNet*}
						&0.932&0.509&0.691&0.678&3.756
						&0.905&0.516&0.786&\textbf{0.645}&\textbf{2.637}
						&0.909&0.400&0.699&0.491&2.529\\			
						&\cellcolor[rgb]{0.7,0.8,1.0}{AViNet\#}
						&\textbf{0.936}&\textbf{0.511}&\textbf{0.694}&0.679&\textbf{3.759}&\textbf{0.906}&\textbf{0.518}&\textbf{0.797}&0.643&2.634&\textbf{0.913}&\textbf{0.405}&\textbf{0.702}&\textbf{0.492}&\textbf{2.535}\\		
						\Xhline{1.0pt}
						\cline{1-17}
%%%%%%%%%%%%%%%%%%%%%%%%%%%%%%%%%%%%%%%%%%%%%%%%%%%%%%%%%%%%%%%%%%%%%%%%%%%%%%%%%%%%%%%%%%%%%%%%%%%%%%%%%%%%%%%%%

						\multirow{2}{*}{Means}&Datasets
						&\multicolumn{5}{c}{{ETMD~\cite{etmd}}}\vline &\multicolumn {5}{c}{{Coutrot1~\cite{coutrot1}}}\vline&\multicolumn{5}{c}{{Coutrot2~\cite{coutrot2}}}\\
						\cline{2-17}
						&Methods &AUC-J$\uparrow $ 	& SIM$\uparrow $ &s-AUC$\uparrow $ & CC$\uparrow $& NSS$\uparrow $ 	&AUC-J$\uparrow $ 	& SIM$\uparrow $ &s-AUC$\uparrow $ & CC$\uparrow $& NSS$\uparrow $&AUC-J$\uparrow $ 	& SIM$\uparrow $ &s-AUC$\uparrow $ & CC$\uparrow $& NSS$\uparrow $  \\ 	 	
						\hline
						\hline

						\multirow{4}{*}{\rotatebox{90}{\tabincell{c}{Un-\\surpervised}}}
						&{ITTI~\cite{ITTI}}
						& 0.661 &0.127&0.582&0.083&0.425
						&0.616& 0.178&0.529& 0.082&0.319
						& 0.694&0.142& 0.530&0.040&0.331\\				
						&{GBVS~\cite{Jonathan-GBVS}}
						& \textbf{0.856} &0.226& 0.613&\textbf{0.299}&\textbf{1.398}
						&\textbf{0.798}&\textbf{0.253}&0.526&\textbf{0.272}& \textbf{1.055}
						&0.819&\textbf{0.189}&0.577&\textbf{0.183}&1.071
						\\							
						&{SBF~\cite{SBF}}
						& 0.805 &\textbf{0.232}&0.641&0.262&1.298
						&0.726&0.187&0.530&0.215&0.789
						&\textbf{0.827}&0.152& 0.583&0.131&\textbf{1.101}
						\\							
						&{AWS-D~\cite{AWS-D}}
						&0.754 &0.161&\textbf{0.664}&0.181&0.907
						&0.729&0.214&\textbf{0.581}&0.207&0.872
						&0.783&0.170&\textbf{0.590}&0.146& 0.842
						\\		
						
						\cline{1-17}
						\Xhline{1.0pt}
						\multirow{6}{*}{\rotatebox{90}{\tabincell{c}{Weakly-\\surpervised}}}
						&{GradCAM++~\cite{Grad-CAM}}
						&0.575 &0.124& 0.157&0.576& 0.736
						&0.704& 0.137&0.537&0.210&0.511
						& 0.733&0.114& 0.567&0.168&0.625
						\\
						&{WSS~\cite{WSS}}
						& 0.854 &0.277& 0.661&0.334&1.650
						&0.772&0.247&0.547&0.233&0.975
						& 0.835&0.208&0.578&0.192&1.178
						\\		
						&{MWS~\cite{MWS}}
						& 0.833&0.237& 0.649&0.293&1.425
						&0.743& 0.231&0.528&0.201& 0.798
						& 0.839&0.188& 0.581&0.168&1.197
						\\				
						\cline{2-17}	
						&{STANet~\cite{AVS-WGT21}}
						&0.908 &0.318 &0.682 &0.448 &2.176
						&0.829 &0.306 &0.542 &0.339 &1.376
						&0.850 &0.247 &0.597 &0.273 &1.475
						\\
						&\cellcolor[rgb]{1,0.8,0.8}{STANet*}
						&0.922 &0.319&0.701&0.464&2.326
						&0.837&0.315&0.550&0.341&\textbf{1.394}
						&0.887&0.264&0.602&0.336&1.915
						\\
						&\cellcolor[rgb]{0.7,0.8,1.0}{STANet\#}
						&\textbf{0.925}&\textbf{0.323}&\textbf{0.704}&\textbf{0.467}&\textbf{2.328}&\textbf{0.840}&\textbf{0.318}&\textbf{0.551}&\textbf{0.346}&1.392&\textbf{0.888}&\textbf{0.266}&\textbf{0.605}&\textbf{0.339}&\textbf{1.921}\\	
						\cline{1-17}
						
						\Xhline{1.0pt}
						\multirow{8}{*}{\rotatebox{90}{\tabincell{c}{Fully-\\surpervised}}}
						&{DeepVS~\cite{Jiang-ECCV18}}
						&  0.904&0.349& 0.686&0.461	&2.480
						&0.830& 0.317&0.561&0.359&1.770
						& 0.925&0.259& 0.646&0.449&3.790
						\\ 	
						&{ACLNet~\cite{ACLNet}}
						&0.915 &0.329&0.675&0.477&2.360
						&0.850& 0.361&0.542&0.425& 1.920
						&0.926&0.322&0.594&0.448& 3.160
						\\ 		
						\cline{2-17}
						&{STAViS~\cite{AVS-STAVIS20}}
						&0.931 &0.425 &0.731 &0.569 &2.940
						&0.868 &0.393 &0.584 &0.472 &2.110
						&0.958 &0.511 &0.710 &0.734 &5.280
						\\ 	
						&\cellcolor[rgb]{1,0.8,0.8}{STAViS*}
						&0.946 &0.454&0.758&0.620&3.406
						&0.864&0.398&0.590&0.487&2.203
						&0.959&0.523&0.731&0.738&5.396\\
						&\cellcolor[rgb]{0.7,0.8,1.0}{STAViS\#}
						&\textbf{0.948}&\textbf{0.457}&0.764&\textbf{0.623}&3.401&0.869&0.395&0.592&0.489&2.210&0.963&0.524&0.732&0.739&5.401\\	
						\cline{2-17}
						&{AViNet~\cite{AVS-AVI21}}
						&0.931 &0.410 &0.740 &0.576 &3.070
						&0.891 &0.427 &0.638 &0.561 &2.710
						&0.953 &0.477 &0.739 &0.738 &5.730
						\\ 		
						&\cellcolor[rgb]{1,0.8,0.8}{AViNet*}
						&0.944&0.447&0.761&0.616&3.437
						&\textbf{0.899}&\textbf{0.431}&0.644&0.573&2.770
						&0.963&0.579&0.742&0.806&\textbf{5.993}\\	
						&\cellcolor[rgb]{0.7,0.8,1.0}{AViNet\#}
						&0.946&0.448&\textbf{0.765}&0.617&\textbf{3.441}&0.894&0.428&\textbf{0.647}&\textbf{0.579}&\textbf{2.774}&\textbf{0.967}&\textbf{0.581}&\textbf{0.746}&\textbf{0.809}&5.990\\				
						\Xhline{1.0pt}
						\cline{1-17}
		\end{tabular}}}}
		\label{tab_CAMP1}
		\vspace{-0.2cm}
	\end{center}
\end{table*}

\section{Quantitative Verifications}
\subsection{Datasets}
There exist six publicly available datasets in our AVSD research field, including DIEM~\cite{diem}, AVAD~\cite{avad}, Coutrot1~\cite{coutrot1}, Coutrot2~\cite{coutrot2}, SumMe~\cite{summe}, and ETMD~\cite{etmd}.
Different from the conventional VSD sets, the eye-fixations in these six sets are collected in the audio-visual environment, while, in the VSD sets, the eye-fixations are simply collected without using any audio information.
We briefly introduce these six sets here, and more details can be found via the links of Table~\ref{tab:Datasets}.
Some qualitative demonstrations can be found in Fig.~\ref{fig:dataset}.

The {DIEM} set consists of 84 film clips, covering 26 films, including commercials, documentaries, game trailers, movie trailers, music videos, and news clips. The video scenes in this set are generally complex with strong background interference.

The {AVAD} set targets at exploring the effects of the highly correlated audio and motion on eye movements. The authors of this set tested the human eyes fixation on 45 video sequences, where these tested sequences are 5 to 10-second video clips containing various scenes, \emph{e.g.}, instrumental playing, dancing, and dialogue.

The {Coutrot} set includes Coutrot1 and Coutrot2 subsets. The dynamic nature scenes in the Coutrot1 set are divided into 4 visual categories: single moving object, multiple moving objects, natural landscapes, and human faces. The Coutort2 set's scenes are all conversations, and it can be found that the fixations are most likely to be located on the speaker's face.

The {SumMe} set contains 25 unstructured videos collected from videos taken by users, whose lengths range from 1 minute to 6 minutes. Since all videos in this set are homemade ones, the corresponding background sounds tend to be very noisy, and most of them are irrelevant to the salient objects, making the audio-visual fusion process very challenging.

The {ETMD} set contains 12 videos, which are all collected from 6 existing Hollywood movies. Each video in this set ranges from 3 to 3.5 minutes, whose contents mainly consist of action scenes and dialogues.

\subsection{Evaluation Metrics}
\label{Metrics}
There are totally five quantitative metrics that have been widely used in the saliency detection field.
Since the objective of measuring the saliency detection performance in an audio-visual environment is almost the same as the conventional saliency detection field, all these five metrics can be directly used here, and we shall briefly introduce them respectively.
These metrics include AUC-Judd (AUC-J), \textbf{sim}ilarity metric (SIM), \textbf{s}huffled \textbf{AUC} (s-AUC), \textbf{n}ormalized \textbf{s}canpath \textbf{s}aliency (NSS), and linear \textbf{c}orrelation \textbf{c}oefficient (CC).

\begin{table*}[!t]
	\caption{Ablation study regarding different AVC classifiers, \emph{e.g.}, L3Net, AVENet, and AVID. The target AVSD model used here is AVINet~\cite{AVS-AVI21}. AVID+\emph{ws} denotes that the AVID-based AVC classifier is trained in a weakly-supervised manner the same as~\cite{AVCLLL-ICCV17}. The bests are highlighted in \textbf{bold} font.}
	\vspace{-0.3cm}
	\begin{center}
		\newcommand{\tabincell}[2]{\begin{tabular}{@{}#1@{}}#2\end{tabular}}
		\renewcommand{\arraystretch}{1.2}{
			\setlength{\tabcolsep}{4.0pt}{
				\resizebox{1.0\linewidth}{!}{
					\begin{tabular}{r||c|ccccc|ccccc|ccccc}
						\Xhline{1.0pt}
						\cline{1-17}
						
						Datasets&
						\multirow{2}{*}{Accuracy}&\multicolumn{5}{c}{{AVAD~\cite{avad}}}\vline &\multicolumn {5}{c}{{DIEM~\cite{diem}}}\vline&\multicolumn{5}{c}{{SumMe~\cite{summe}}}\\
						%\cline{1-1}
						\cline{3-17}
						Methods &&AUC-J$\uparrow $ 	& SIM$\uparrow $ &s-AUC$\uparrow $ & CC$\uparrow $& NSS$\uparrow $ 	&AUC-J$\uparrow $ 	& SIM$\uparrow $ &s-AUC$\uparrow $ & CC$\uparrow $& NSS$\uparrow $&AUC-J$\uparrow $ 	& SIM$\uparrow $ &s-AUC$\uparrow $ & CC$\uparrow $& NSS$\uparrow $  \\ 	 	
						\hline
						\hline																							
						
						L3Net (\emph{ours})
						&82.15\%&0.928&0.505&0.682&0.674&3.750
						&0.902&0.514&0.768&0.639&2.605
						&0.902&0.377&0.698&0.483&2.489\\					
						AVENet (\emph{ours})
						&85.64\%&0.929&0.507&0.685&0.677&3.753
						&0.903&0.511&0.779&0.642&2.617
						&0.907&0.392&\textbf{0.699}&0.488&2.510
						\\
						\cline{1-17}	
						AVID + \emph{ws}~\cite{AVCCL1-ARXIV21}
						&85.82\%&0.915&0.487&0.667&0.658&3.652&0.891&0.493&0.755&0.628&2.589&0.894&0.369&0.680&0.473&2.474\\							
						AVID (\emph{ours})
						&\textbf{87.59}\%&\textbf{0.932}&\textbf{0.509}&\textbf{0.691}&\textbf{0.678}&\textbf{3.756}
						&\textbf{0.905}&\textbf{0.516}&\textbf{0.786}&\textbf{0.645}&\textbf{2.637}
						&\textbf{0.909}&\textbf{0.400}&\textbf{0.699}&\textbf{0.491}&\textbf{2.529}
						\\						
						
						\Xhline{1.0pt}
						\cline{1-17}				
		\end{tabular}}}}
		\label{ablation}
		\vspace{-0.4cm}
	\end{center}
\end{table*}

CC is a method to measure the linear correlation between the prediction saliency (S) and the ground truth (GT), which can be formulated as:
\begin{equation}
\begin{aligned}
&{\rm CC(S,GT)}=\frac{cov({\rm S},{\rm GT})}{\sigma({\rm S})\cdot\sigma({\rm GT})},
\end{aligned}
\label{eq:CC}
\end{equation}
where $cov$ denotes the covariance, and $\sigma$ is the standard deviation.

SIM measures the similarity between two distributions. Given S and GT as input, SIM first normalizes them respectively, then measures the minimum values pixel-by-pixel (denoted by $i$). This process can be detailed as:
\begin{equation}
		{\rm SIM}= \sum_{i} min\Big\{\mathcal{Z}({\rm S})_i, \mathcal{Z}({\rm GT})_i\Big\},
	\label{eq:SIM}
\end{equation}
where $\mathcal{Z}$ and $min$ respectively denote the normalization operation and minimum operation.

AUC measures the area under the \textbf{r}eceiver \textbf{o}perating \textbf{c}haracteristic (ROC) curve, which has been widely used to evaluate the maps by saliency models.
Given an image and its ground-truth eye fixations, the fixated points are regarded as the positive set, and others are regarded as the negative set.
Then, the computed saliency map is binarized into salient regions and non-salient regions by using a hard threshold.
The AUC-Judd (AUC-J) computes two items: 1) the true positives from all the saliency map values above a threshold at fixated pixels and 2) the false positive rate as the total saliency map values above a threshold at non-fixated pixels.
The s-AUC samples the negatives from fixated locations of other images/frames. This sampling scheme can be greatly influenced by center bias and border cuts.

The NSS is designed to evaluate a saliency map over fixation locations. Given a saliency map S and a binary fixation map GT, NSS is defined as:
\begin{equation}
	\begin{aligned}
		{\rm NSS}=\frac{1}{\rm M}\sum_i^{\rm M} \hat{\rm S}_i\cdot {\rm GT}_i,\ \ {\rm M} \gets \sum_{i}{\rm GT}_i,\ \ \hat{\rm S} \gets \frac{{\rm S}-\mu}{\sigma},
	\end{aligned}
	\label{eq:NSS}
\end{equation}
where $\mu$ and $\sigma$ are the mean and standard deviation of the predicted saliency map.
This metric is calculated by taking the mean of scores assigned by the unit normalized saliency map (with zero mean and unit standard deviation) at human eye fixations.

\subsection{Quantitative Evidences towards the Effectiveness of the proposed AVC Classifier}
As we have mentioned before, our approach is generic, and is compatible with almost all existing bi-stream SOTA AVSD models.
By using a few lines of code modification, the proposed AVC classifier can be intergraded into the target model.
To verify this issue, we have tried to deploy our AVC classifier into 3-top tier SOTA AVSD models, including STANet~\cite{AVS-WGT21}, STAVIS~\cite{AVS-STAVIS20}, and AVINet~\cite{AVS-AVI21}.
In fact, we shall incorporate our AVC classifier into more SOTA models, yet, in the AVSD research field, most of the existing papers haven't released their codes.
Also, \emph{w.r.t.} the model training, we follow the widely-used training/testing split~\cite{AVS-STAVIS20} over all 6 datasets.
To demonstrate the superiority of our approach, we have compared the upgraded versions of the three target models (denoted by *) with 12 other SOTA methods, including 4 unsupervised methods, 4 weakly-supervised methods, and 4 fully-supervised methods.
For a fair comparison, we use either the code implementations with default parameter settings or saliency maps provided by the authors.
Specifically, for others without codes, we simply refer to the numeric results reported in the papers.

As is shown in Table~\ref{tab_CAMP1}, all three upgraded target models (denoted by * highlighted by PINK color) can achieve persistent performance improvements.
For example, our method can make an average of 1.9\%, 1.5\%, and 2.7\% performance improvement generally of STANet, STAVIS, and AVINet respectively in terms of the AUC-J metric on six widely-used benchmark datasets.
Also, the promoted model STANet* outperforms all weakly-supervised methods significantly and AVINet* performs the best among all fully-supervised methods.
The main reason is that the AVSD benchmark datasets equipped with the newly proposed AVC classifier can filter out the unrelated audio-visual pairs, so that the side-effects from those mismatched audio-visual fragments can be avoided.

To further investigate the importance of our key idea, \emph{i.e.}, the audio-visual consistency really matters when performing AVSD, we have additionally removed the proposed AVC classifier from the upgraded target AVSD models.
Instead, we directly use the original versions, yet their outputs are ``manually reformulated'' according to our newly provided AVC labels (\emph{i.e.}, \textbf{Lc} in Fig.~\ref{fig:pipeline}).
That is, the target model's output will be derived directly by using either AV or V, and this process can be formulated as:
\begin{equation}
\begin{aligned}
{\rm OUTPUT}&\gets {\textbf{Lc}} \cdot {\rm Fuse}({\rm AV},{\rm V}) + (1-{\textbf{Lc}})\cdot {\rm V}. \\
\end{aligned}
\label{eq:score2}
\end{equation}
where all symbols are identical to Eq.~\ref{eq:score}, and the major difference is that the $\tilde{\textbf{Lc}}$ has been replaced by \textbf{Lc}.
Actually, OUTPUT from Eq.~\ref{eq:score2} is in ideal situation, which tends to persistently outperform that from the upgraded version powered by the AVC classifier (\emph{i.e.}, Eq.~\ref{eq:score}).
The main reason is quite clear, and erroneous binary predictions can not be completely avoided by our AVC classifier.
The quantitative results of these ideal versions have been marked by $\#$ with BLUE background color, and the detailed results can be found in Table~\ref{tab_CAMP1}.

Further, as mentioned above, the classification accuracy of the AVC classifier will affect AVSD performance slightly. Thus, we have tested three AVC classifiers to verify this issue, \emph{i.e.}, L3Net~\cite{AVCLLL-ICCV17}, AVENet~\cite{AVE-ECCV18}, and AVID~\cite{AVCCL1-ARXIV21}.
Among them, the AVID is our default setting, and the other two classifiers can simply be used to replace the AVID in our method as shown in Fig.~\ref{fig:pipeline} (b).
That is, in each experiment, we only replace the target AVSD model's AVC classifier with either L3Net, AVENet, or AVID.
The experimental results have been shown in Table~\ref{ablation}.

Actually, the influence of the classification result is based on the amount of corresponding audio-visual pairs, \emph{e.g.}, the more the corresponding audio-visual pairs are, the better the performance of the target models obtain; otherwise, the target models will degenerate into the original versions.
According to the results, the AVID-based AVC classifier has achieved the best accuracy (\emph{i.e.}, 87.59\%), and thus, as expected, the corresponding AVSD performance outperforms others.
In short, the higher the performance of the AVC classifier is, the better the performance of the target models obtains.

\section{Conclusions and Future Work}
In this paper, we present the first comprehensive review covering both topics ranging from saliency detection to audio-visual fusion.
Based on this extensive review, we also provided a deep insight into the audio-visual saliency detection task, and reached our new claim about the importance of an AVSD model to be audio-visual consistency aware (AVC-aware).
We have also devised a generic method to convert the existing AVC-unaware SOTA AVSD models to be AVC-aware, and the key is the newly proposed AVC classifier, which, as a plug-in, controls the data flow of the bi-steam target AVSD mode to avoid side-effects caused by mismatched audio-visual training fragments.
Specifically, to train the proposed AVC classifier, we have newly labeled all existing publicly available AVSD datasets, equipping them with AVC labels.
Lastly, we have conducted extensive experiments to verify the effectiveness of our claim.
Hoping this review could draw more research attentions to the AVSD research field, and the newly claimed AVC-aware issue could inspire future works in performance improvement.

Specifically, although audio-visual-based saliency detection has made notable progress over the past several decades, there is still significant room for improvement, \emph{i.e.}, the AVSD model can only obtain limited performance. Thus, in the near future, we are particularly interested in further designing a more reasonable AVC classifier to improve the performance of audio-visual correspondence.

\vspace{-0.1cm}
\bibliographystyle{IEEEtran}
\bibliography{TIP_reference}

\end{document}